\documentclass{ieeeaccess}
   \usepackage{cite}
\usepackage{amsmath,amssymb,amsfonts}
\usepackage{algorithmic}
\usepackage{graphicx}
\graphicspath{ {./images/} }
\usepackage{CJK}
\usepackage{textcomp}
\usepackage{comment}
\usepackage{hyperref}
\usepackage{flushend}
\usepackage{xcolor, soul}
\sethlcolor{white}
\usepackage{array}
\usepackage{multirow}
\usepackage{mathabx}
\usepackage{algorithm}
\usepackage{algorithmic}

\newcommand\MyBox[2]{
 \fbox{\lower0.75cm
  \vbox to 1cm{\vfil
   \hbox to 1cm{\hfil\parbox{1.4cm}{#1\\#2}\hfil}
   \vfil}%
 }%
}

\def\BibTeX{{\rm B\kern-.05em{\sc i\kern-.025em b}\kern-.08em
 T\kern-.1667em\lower.7ex\hbox{E}\kern-.125emX}}
\begin{document}
\history{Date of publication xxxx 00, 0000, date of current version xxxx 00, 0000.}
\doi{10.1109/ACCESS.2022.0092316}

\title{Event Detection and Classification for Long Range Sensing of Elephants Using Seismic Signals}

\author{\uppercase{Jaliya L. Wijayaraja}\authorrefmark{1,*},\IEEEmembership{Member, IEEE},
\uppercase{Janaka L. Wijekoon}\authorrefmark{2,3,*},\IEEEmembership{Senior Member, IEEE}, \uppercase{Malitha Wijesundara}\authorrefmark{1.*},\IEEEmembership{Member, IEEE}
}

\address[1]{Sri Lanka Institute of Information Technology, New Kandy Rd, Malabe 10115, Sri Lanka}
\address [2]{Victorian Institute of Technology, Adelaide Campus, South Australia}
\address[3]{Department of System Design Engineering, Keio University, Yokohama, Japan}

\markboth
{Author \headeretal: Preparation of Papers for IEEE TRANSACTIONS and JOURNALS}
{Author \headeretal: Preparation of Papers for IEEE TRANSACTIONS and JOURNALS}

\corresp{Corresponding author: Jaliya L. Wijayaraja (e-mail: jaliya.wijayaraja@gmail.com).}

\begin{abstract}
Detecting elephants through seismic signals is an emerging research topic aimed at developing solutions for Human-Elephant Conflict (HEC). Despite the promising results, such solutions heavily rely on manual classification of elephant footfalls, which limits their applicability for real-time classification in natural settings. To address this limitation and build on our previous work, this study introduces a classification framework targeting resource-constrained implementations, prioritizing both accuracy and computational efficiency. As part of this framework, a novel event detection technique named Contextually Customized Windowing (CCW), tailored specifically for detecting elephant footfalls, was introduced, and evaluations were conducted by comparing it with the Short-Term Average/Long-Term Average (STA/LTA) method. The yielded results show that the maximum validated detection range was 155.6 m in controlled conditions and 140 m in natural environments. Elephant footfall classification using Support Vector Machine (SVM) with a Radial Basis Function (RBF) kernel demonstrated superior performance across multiple settings, achieving an accuracy of 99\% in controlled environments, 73\% in natural elephant habitats, and 70\% in HEC-prone human habitats, the most challenging scenario. Furthermore, feature impact analysis using explainable AI identified the number of Zero Crossings and Dynamic Time Warping (DTW) Alignment Cost as the most influential factors in all experiments, while Predominant Frequency exhibited significant influence in controlled settings.

\end{abstract}


\titlepgskip=-15pt

\maketitle

\section{Introduction}
\label{sec:introduction}

The human-elephant conflict (HEC) poses significant challenges in wildlife management, socio-economic stability, and environmental conservation across a range of countries \cite{shaffer2019human}. Notably, Sri Lanka faces acute challenges from HEC due to its high density of elephant populations \cite{thant2021pattern,santiapillai2010assessment}, reporting the highest annual elephant fatalities and the second-highest human casualties globally \cite{prakash2020human}. A promising approach to managing HEC involves the early detection of elephants using non-invasive methods \cite{wijayaraja2024towards}. Among these, seismic signal-based detection demonstrates a significant potential \cite{wood2005using}.

\hl{Seismic signals generated by elephants propagate over long distances with minimal interference, \mbox{\cite{arnason2002properties, wijayaraja2024towards, szenicer2022seismic}},  while retaining distinctive characteristics that enable their differentiation from other environmental seismic disturbances. \mbox{\cite{wood2005using,wijayaraja2024towards}} Notably, seismic-based detection offers distinct advantages over other non-invasive methods, such as visual, \mbox{\cite{premarathna2020mage,premarathna2020cnn,ravikumar2020layered, pemasinghe2023development, jothibasu2023improvement, mondal2023design, moorthy2025deep, rahuman2025using, kommalapati2024drone, sawal975mobilenetv2}} acoustic, \mbox{\cite{zeppelzauer2013acoustic, bjorck2019automatic, prince2014surveillance, arya2016design, ranasinghe2023enhanced, ramasubramanian2022averting, deowan2022warning, dewmini2025elephant, geldenhuys2025learning }} and infrasonic, \mbox{\cite{prince2014surveillance, sayakkara2017eloc}} detection. Specifically, seismic detection has demonstrated strong potential for reliable, long-range identification of elephants, with robust and sustainable performance under tropical conditions \mbox{\cite{wijayaraja2024towards}}. Therefore, we previously developed an embedded instrumentation system featuring a novel geophone-sensor interface, specifically tailored for detecting elephant-generated seismic signals over extended distances \mbox{\cite{wijayaraja2024towards}}.}

The introduced instrumentation system in \cite{wijayaraja2024towards} demonstrate a capability to detect elephants at a maximum distance of 155.6 m with an overall accuracy of 99.5\%. This system was specifically focused on Sri Lankan elephants (Elephas maximus maximus) a distinct subspecies of Asian elephants  \cite{fernando2011current,nozawa1990genetic}, and was tested under controlled, partially-controlled, and natural environments with minimal noise. However, the classification process relied heavily on manual intervention. Researchers manually inspected each recording, compared with reference videos, and split and segmented the footfall events for further analysis. Subsequently, these events were categorized into classes (elephant footfall, human footfall, noise from a motorcycle) based on their predominant frequency, utilizing a Decision Tree (DT) model.

While manual splitting is effective for research applications, developing a field-implementable early detection solution for HEC necessitates automating the entire process from signal acquisition to classification. Hence, automating the process by utilizing event detection and extraction algorithms becomes imperative. Given the natural environments often prone to significant ambient noise, the existing method in \cite{wijayaraja2024towards}, which relies on a single feature for event detection, may fail in complex natural environments despite showing satisfactory results in controlled settings. Therefore, it is vital to explore additional features and develop more sophisticated classification models tailored to Sri Lankan elephants and resilient to the natural environmental conditions prevalent in Sri Lanka.

To this end, this study first introduces a dedicated filter designed to suppress seismic noise commonly present in natural environments. Subsequently, a novel event detection method termed the context-customized window (CCW), which is specifically tailored for elephant footfall signals, is proposed. The CCW is then compared with two other existing event detection methods, namely, Short-Term Average/Long-Term Average (STA/LTA) and the modified Energy Ratio (MER), which are commonly used in seismology applications \cite{lee2017improved}. These event detection methods will be evaluated based on their accuracy and execution time. Additionally, this study discusses nine distinct features, categorized into temporal, spectral, pattern-matching, and statistical types, that were extracted from the detected events for further classification. 

This study assesses two classification models for their ability to classify the extracted features and identify signals generated by elephant footfalls. A Support Vector Machine (SVM) was employed as a low-complexity model suitable for resource-constrained environments \cite{madhu2021comparative,anguita2007hardware}, such as embedded systems. Same datasets were also used to train an Artificial Neural Network (ANN) \cite{madhu2021comparative} as a reference model. Both models were trained and tested using datasets form various conditions, including controlled environments and natural settings resembling HEC-prone areas. Finally, this study analyzes the impact of each feature on the classification accuracy of elephants in both natural and controlled conditions to guide future improvements.

\section{Background Survey}
\label{sec:Background}
Existing studies on event detection and classification using seismic signals have focused mainly on specific domains \cite{zhou2022adaptive,hu2024intrusion,hettigoda2020surveillance,wijayaraja2024towards}. Notably, studies such as \cite{cyriac2022seismic,mukhopadhyay2017detection,choudhary2021seismic,hettigoda2020surveillance} present successful approaches utilizing seismic signal detection and classification in recognizing human activities, confirming that seismic signals are a viable means of footfall classification and identification. Among the aforementioned studies, \cite{mukhopadhyay2017detection} proposed a strategy specifically tailored for embedded systems, integrating event detection, feature extraction, and event classification.

The study in \cite{mukhopadhyay2017detection} highlights the importance of event detection, particularly when event detection is employed to identify the presence and location of significant seismic events within a signal. Specifically, \cite{mukhopadhyay2017detection} introduces event detection methods including amplitude thresholding, kurtosis-based event detection, STA/LTA, and MER which are utilized before classification. Similarly, \cite{anchal2017predicting} proposed an advanced adaptive approach for event detection using STA/LTA method. However, similar to \cite{mukhopadhyay2017detection}, the scope of  \cite{anchal2017predicting} also remains limited to human footfall. 

Nevertheless, the study \cite{parihar2021seismic}, which is closely aligned with one of the objectives of this study, introduced three methods to detect elephant footfall events: the STA/LTA algorithm, continuous wavelet transforms (CWT), and amplitude spectrum of the Fourier transform (ASFT). However, \cite{parihar2021seismic} is limited to event detection; it does not extend to classifying the detected events to differentiate elephants’ seismic signals from other seismic activities. 

\hl{Conversely, \mbox{\cite{goderik2024seismic}} employed kurtosis analysis to distinguish seismic events from ambient noise, followed by classification that achieved 89\% accuracy in identifying elephants using feature thresholding based on the standard deviation of the signal, frequency peak, spectral centroid, and frequency distribution. Despite demonstrating the potential importance of event detection methods, the study detected only 54\% of footsteps and was limited to a range of 8 to 30 meters, focusing on African elephants.}

\hl{Despite the lack of event detection, an essential component for achieving practical HEC solutions, several studies successfully concluded that various distinctive characteristics of seismic waves generated by elephants can be used to classify and differentiate them from seismic signals generated by other animals or human activities \mbox{\cite{wood2005using,szenicer2022seismic}}. Such studies are summarized in Table \mbox{\ref{tab:classification_lit_summery}}}.

\begin{table*}[!t]
\centering
\caption{Summary of literature on classification and detection of elephants using seismic signals }
\label{tab:classification_lit_summery}
\begin{tabular}{|c|p{1.4 cm}|p{1 cm}|p{1 cm}|p{1.2 cm}|c|p{4.2 cm}|p{1.2 cm}|}
\hline
\textbf{Ref.}&\textbf{Acquisition Method}&\textbf{Elephant Type}&\textbf{Event Type}&\textbf{Detection Range}&\textbf{Classification Method}&\textbf{Features Employed}&\textbf{Reported Accuracy}\\ \hline
{\cite{wood2005using}}&So&African&L&$\sim$100 m&Discriminate analysis&Correlation coefficient of power spectra&82\%\\ \hline
{\cite{nakandala2014detecting}}&So&Asian&L&CL&MLP&Signal recordings&50\%\\ \hline
{\cite{szenicer2022seismic}}&Se&African&V/L&150 m&CNN&Spectrogram&73-90\%\\ \hline
{\cite{fernando2020gaja}}&Em&Asian&L&3-5 m&ANN&Signal recordings&100\%\\ \hline
{\cite{thilakarathne2022prototype}}&Em&NA&L&NA&Logistic regression&Signal recordings&93\%\\ \hline
{\cite{parihar2022variational}}&Em&Asian&L&NA&SVM&Ref Table 2&73\%\\ \hline
{\hl{\mbox{\cite{vidunath2024identification}}}}&Em&Asian&V&NA&Ridge classifier**&MFCC**&97.05\%\\ \hline
{\cite{chandan2024gajgamini}}&Em&Asian&L&1-25 m&CNN**&Signal recordings&98.03\%\\ \hline
{\hl{\mbox{\cite{deb2025implementation}}}}&\hl{Em}&\hl{Asian}&\hl{L}&\hl{2 m}&\hl{Feature thresholding}&\hl{Signal frequency and amplitude, Cumulative volume of vibration}&\hl{80\%}\\ \hline
{\hl{\mbox{\cite{parihar2025hybrid}}}}&\hl{Em}&\hl{Asian}&\hl{L}&\hl{NA}&\hl{CNN-BiLSTM**}&\hl{29 multi-domain features \footnote{Features are not explicitly mentioned}}&\hl{77.89\%}\\ \hline
{\hl{\mbox{\cite{steinmann2025decoding}}}}&\hl{Se}&\hl{African}&\hl{L}&\hl{150 m}&\hl{Support vector classifier}&\hl{Scattering transform features extracted from seismogram}&\hl{77\%}\\ \hline
{\hl{\mbox{\cite{tv2024seismic}}}}&\hl{Em}&\hl{NA}&\hl{L}&\hl{NA}&\hl{ Amplitude thresholding}&\hl{Average signal amplitude}&\hl{NA}\\ \hline
\end{tabular}

\vspace{1mm} 
\small
\footnotesize
\begin{flushleft}
\textbf{Note:} So – Sound card; Se – Seismometers; Em – Embedded Systems; L – Locomotion; V – Vocalization; CL – Close to path; MLP – Multi Layer Perceptron; CNN – Convolutional Neural Network; ANN – Artificial Neural Network; MFCC – Mel Frequency Cepstral Coefficients; SVM – Support Vector Machine; SVC – Support Vector Classifier; LSTM – Long Short-Term Memory. \textbf{**} This study trialed multiple methods; the given method is the best among them.
\end{flushleft}
\end{table*}

\begin{table*}[!t]
\centering
\caption{ Features utilized in related studies for identifying footfall seismic signals of elephants}
\label{tab:featutres_lit}
\begin{tabular}{|c|l|}
\hline
\textbf{Feature Type} & \textbf{Feature} \\ \hline
Temporal & Energy of Signal, Zero Crossing Rate, Skewness, Kurtosis, Shannon Entropy \\ \hline
Spectral & Energy of Frequency Bands, Spectral Cepstral \\ \hline
Cepstral & Log energy of MFCC \\ \hline
\end{tabular}
\end{table*}

\hl{As detailed in the Table \mbox{\ref {tab:classification_lit_summery}},} an early study conducted in 2005 by Wood et al. \cite{wood2005using} employed a discriminant analysis to achieve a maximum accuracy of 82\% in detecting seismic signals generated by elephant footfalls, with a relatively higher detection range of approximately 100 meters for African elephants in their natural habitats. This study used a sound card to collect seismic signals, and a correlation coefficient was used for event classification. Later, the same setup was followed by \cite{nakandala2014detecting} for Sri Lankan elephants employing the multilayer perception (MPL) method that directly used the signal recordings. Although the experiments were conducted in a controlled environment in a very close area, this study reported the lowest accuracy among comparable studies. According to the authors, the use of a generic sound card instead of a high-end sound card (that is, the Digigram vx pocket v2 used in \cite{wood2005using} hindered the performance. However, it is a notable fact that reliance on sound cards — a computer peripheral designed for indoor use, obviously limits the feasibility as field-deployable solutions for HEC due to limited durability in outdoor tropical terrains.

In a similar study, \cite{szenicer2022seismic} employed standard seismic data acquisition methods (i.e., 22 6TD Guralp seismometers) to record the seismic signal of African elephants in their natural habitats. This study achieved 80–90\% accuracy in classifying elephant footfalls at a range of 100 m by using spectrograms of elephant footfall recordings together with a Convolutional Neural Network (CNN). However, accuracy dropped to 73\% when the range was increased to 150 m, highlighting the critical impact of range on the detection accuracy of the classifier. Furthermore, the study highlighted that the higher detection ranges were obtained under the limitation of using trained and tested data from the same seismometer and from matching environmental conditions from different locations, which restricts the classifier’s usability as an applicable solution for HEC. \hl{Using the same dataset, \mbox{\cite{steinmann2025decoding}} improved the accuracy to 77\% at 150 meters by employing a Support Vector Classifier (SVC). Nevertheless, the dependence on high-cost seismometers limits their feasibility and scalability of a cost-effective solution for mitigating HEC.}

The recent studies highlight the fact that the generic embedded systems are also emerging as a feasible approach for acquiring seismic signals and classifying elephants, offering customizability as a practical solution for HEC \cite{wijayaraja2024towards}. Notably, authors from 'Gaja-Mithuru' reported that 100\% of classification accuracy was achieved by using an artificial neural network, with sensors deployed at close distances of 1–3 m \cite{fernando2020gaja}. However, the model in \cite{fernando2020gaja} was trained to distinguish and categorize walking and running, and no effort was made to distinguish the elephants’ footfalls from other ambient signals.

Similarly, \cite{chandan2024gajgamini} reported the highest accuracy of 98.03\% using their own CNN model named ‘GajGamini’ while comparing it with several other classification methods including Random Forest, Support Vector Machine (SVM) with a Radial Basis Function (RBF) kernel, K-Nearest Neighbors (KNN), DT, MLP, and a CNN model introduced in \cite{szenicer2022seismic} named ‘SeisSavanna.’ Both these methods used signal recordings without any feature extraction for the classification. While conventional machine learning models typically exhibit lower computational complexity \cite{mukhopadhyay2017detection}, this study in \cite{chandan2024gajgamini} highlights that, in real-time implementation, the proposed CNN model outperformed conventional machine learning models by eliminating the need for explicit feature extraction. \hl{This, in turn, underscores the importance of optimizing conventional machine learning methods through the elimination of less significant features to maintain computational efficiency.}

Additionally, \cite{thilakarathne2022prototype} achieved a 93\% accuracy of identifying elephants using logistic regression directly on the signal recordings. \hl{Parihar et al. \mbox{\cite{parihar2022variational}}, on the other hand, suggested using nine different features (presented in Table \mbox{\ref{tab:featutres_lit}}) to classify the elephants instead of relying on raw signal recordings. These features represent multiple domains of the signal, including temporal, spectral, and cepstral characteristics. Subsequently, these features were used for the classification of seismic signals using an SVM classifier.} However, the results revealed that the classification achieved only 73\% accuracy under natural environments. \hl{Building on the same dataset, \mbox{\cite{parihar2025hybrid}} improved the accuracy by approximately 5\% by employing a CNN-BiLSTM model with 29 multi-domain features.}

\hl{Vidunath et al. in \mbox{\cite{vidunath2024identification}}, in a recent study}, used vocalization for elephant classification instead of using footfalls. They employed a combination of feature extraction methods to extract spectral features, along with various classifiers. Specifically, Mel-frequency cepstral coefficients (MFCC) were classified with a Ridge Classifier, SVM, and Logistic Regression; Hjorth parameters were classified using a Decision Tree (DT), AdaBoost Classifier, and Random Forest Classifier; and spectral energy distribution was classified with a Light Gradient Boosting Machine (LGBM), Gradient Boosting Classifier, and AdaBoost Classifier. The best performance achieved was 97.05\% with MFCC and a Ridge Classifier.
\hl{Moreover, another publication \mbox{\cite{costa2023detection}} from the same study stated that only seven rumbles were recorded during several hours of observation}. Additionally, vocalizations were found to propagate over shorter distances compared to footfall signals \cite{o2000seismic}, highlighting the reduced potential of vocalizations for long-range and accurate classification compared to locomotion.

\hl{Interestingly, recent studies in this domain suggest more integrated approaches that combine real-time seismic signal classification with remedial interventions \mbox{\cite{deb2025implementation}, \cite{tv2024seismic}}. Both studies relied on basic threshold-based detection techniques, and \mbox{\cite{deb2025implementation}} did not focus on field-level detection accuracy or operational range, whereas \mbox{\cite{deb2025implementation}} employed a two-sensor string-based configuration instead of conventional geophones, reporting 80\% accuracy with a limited detection range of 2 meters. Most importantly, \mbox{\cite{deb2025implementation}} also identifies critical limitations, including misclassification due to overlapping frequency patterns from non-target animal groups and performance degradation caused by topographical conditions. These findings highlight the necessity for more advanced classification methodologies to enable accurate, long-range detection in diverse environmental contexts.}

Apart from our previous work in \cite{wijayaraja2024towards}, the handful of studies conducted using generic embedded systems \cite{fernando2020gaja,thilakarathne2022prototype,parihar2022variational,vidunath2024identification,chandan2024gajgamini} showed a very limited range in detecting elephant footfalls. Although the reported accuracies are satisfactory, most of these studies derived their results from data collected in controlled environments, as mentioned in \cite{fernando2020gaja,parihar2022variational,vidunath2024identification,chandan2024gajgamini}, hindering the implementations in natural HEC prone habitats. Additionally, the study in \cite{thilakarathne2022prototype} utilized secondary data for which the species of the elephants under study and the nature of the environment were not confirmed. Still, related comprehensive studies have concluded that the environmental conditions of the training dataset are critical to the accuracy of the final implementation \cite{szenicer2022seismic}. Therefore, it is essential to conduct further experiments to train and analyze the performance of the classifiers in natural habitats of elephants where HEC is prone to occur, particularly for Sri Lankan elephants.

In summary, there is significant potential to automate the process outlined in our previous work \cite{wijayaraja2024towards} by utilizing event detection, event extraction, and feature extraction as presented in related studies. Also, the available literature reveals that there is a gap in developing a classifier that is resilient to natural environments prone to HEC, particularly for Sri Lankan elephants, while enabling long-range detection. Moreover, existing literature has concluded that feature extraction  is computationally expensive for classifying and identifying elephants using seismic signals \cite{chandan2024gajgamini}. Thus, further analysis is required to identify the most impactful features to optimize conventional machine learning models, particularly for resource-constrained environments such as embedded systems.

\section{Methodology}
\label{sec:Methodology}
\begin{figure*}[!t]
 \centering
 \includegraphics[width=0.55\textwidth]{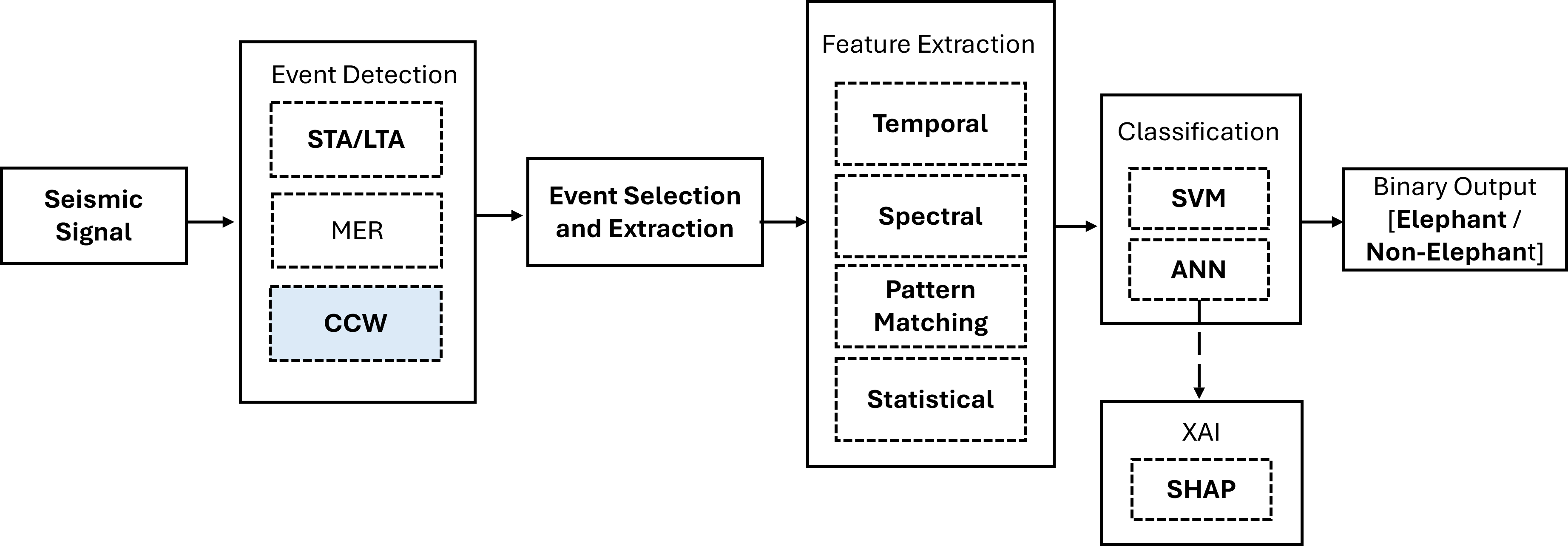}
 \caption{\textbf{ Overview of the proposed seismic signal classification pipeline.}}
 \label{fig:overallstructure}
\end{figure*}

This study introduces a systematic process for classifying and identifying elephant footfalls, comprising event detection, event extraction, feature extraction, and classification. Subsequently, explainable AI (XAI) techniques were employed to analyze the contribution of each feature to classification accuracy. \hl{This process, illustrated as a pipeline in Fig.   \mbox{\ref{fig:overallstructure}}, is discussed in detail in the subsequent sections.}

\subsection{Event Detection}

Accurate detection of elephant footfall events, while effectively suppressing seismic noise, is essential for the long-term continuous operation of monitoring systems. The use of computationally intensive algorithms for feature extraction and classification often results in inefficient resource utilization, particularly in embedded systems with limited computational capabilities. Our previous study \cite{wijayaraja2024towards} highlighted instances where the instrumentation system generated low-amplitude noise patterns without meaningful events, typically occurring in the absence of  elephant or other significant activity within the monitored area. Furthermore, as noted by \cite{parihar2022variational}, even in the presence of elephants, their natural behavior characterized by intermittent walking interspersed with pauses for feeding or social interactions, results in substantial temporal gaps between footfall events. Similar behavioral patterns were observed in our field studies conducted in Digampathaha, Sri Lanka \cite{wijayaraja2024towards}. Consequently, this study prioritizes the initial detection and extraction of potential elephant footfall events from seismic signals, while minimizing the noise.

In this study, three distinct event detection methods were implemented for comparison and benchmarking: STA/LTA, MER, and CCW. Among these, CCW is a novel approach introduced specifically in this study, while STA/LTA and MER are well-established techniques commonly used in seismological research \cite{lee2017improved,akram2012adaptive}. To establish a baseline for event detection, each scenario was visually inspected using synchronized video recordings from field studies, with relevant events manually identified and annotated. This manually annotated dataset is considered as the ground truth for the evaluations.

Given the importance of accurately defining the duration of seismic events for precise event detection, \hl{ this study manually and systematically  selected 50 samples of elephant footstep events based on waveform morphology} from seismic signal recordings obtained during field studies\footnote{These samples were selected from Dt\_Pinnawala\_Loc2}. The average duration of the seismic signals generated by a single elephant footfall was then calculated and designated as the reference length of a seismic event for Sri Lankan elephants, denoted as $L_{\text{Elp}}$. The minimum and maximum marginal durations of these samples were identified and labeled as $L_{\text{Elp\_Min}}$ and $L_{\text{Elp\_Max}}$, respectively.

\subsubsection{short-term average/long-term average (STA/LTA)}
The STA/LTA method calculates the ratio of average energy in a short-time window to that in a long-time window \cite{mukhopadhyay2017detection, trnkoczy2009understanding}. The short-time window precedes the long trailing window. According to \cite{mukhopadhyay2017detection}, the energy ratio is calculated using (\ref{eq:stalta}) where $er[n]$ represents the energy ratio at testing index $n$. $s$ and $l$ denote the length of the short and long-time windows. Here, $x$ represents the signal function. 

\begin{equation}\label{eq:stalta}
er[n] = \frac{\frac{1}{s} \sum_{n_1 = l}^{l+s} \left( x[n_1] \right)^2}{\frac{1}{l} \sum_{n_2 = n}^{n+l} \left( x[n_2] \right)^2}
\end{equation}

\begin{figure}[!t]
 \centering
 \includegraphics[width=0.4\textwidth]{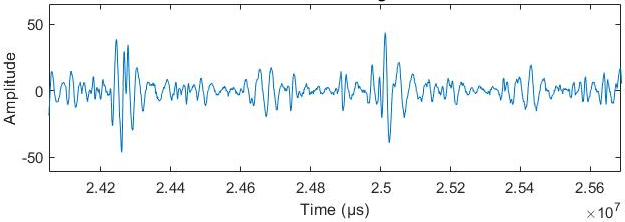}
 \caption{\textbf{Seismic signal of elephants observed in our previous study \cite{wijayaraja2024towards}}}
 \label{fig:FootfallFormPreviousStudy}
\end{figure}

According to general guidelines, the length of the short-time window is typically chosen to be shorter than the shortest expected event  \cite{trnkoczy2009understanding}. Additionally, it is recommended that the length of the long time window be 5 to 10 times that of the short time window \cite{han2009time,mukhopadhyay2017detection}.
Consequently, a ratio of 5 was approximated based on observations of the elephant footfall signal pattern from our previous study in \cite{wijayaraja2024towards},\hl{ as illustrated in Fig. \mbox{\ref{fig:FootfallFormPreviousStudy}}, which captures multiple consecutive footfalls generated by an elephant walking in a straight path with a typical gait.} 

\subsubsection{Modified Energy Ratio (MER)}
The Modified Energy Ratio, an improved version of the energy ratio, enhances event detection by addressing challenges posed by varying signal amplitudes and background noise. According to \cite{mukhopadhyay2017detection}, the energy ratio, $er[i]$, is defined at testing index $i$, where $X$ represents the signal function and $L$ denotes the window length in  (\ref{eq:energy_ratio}). Equation (\ref{eq:modified_energy_ratio}) defines the modified energy ratio at the test index $i$ as $mer[i]$, extending the energy ratio by incorporating the instantaneous amplitude of the signal.

\begin{equation}\label{eq:energy_ratio}
er[i] = \frac{ \sum_{n_1 = i}^{i+L} \left( X[n_1] \right)^2}{ \sum_{n_2 = i-L}^{i} \left( X[n_2] \right)^2}
\end{equation}

\begin{equation}\label{eq:modified_energy_ratio}
mer[i] = \left[ er[i] \cdot \text{abs}(X[i]) \right]^3
\end{equation}

\subsubsection{Contextually Customized Window (CCW)}

Compared to conventional event detection methods used in seismology, such as STA/LTA and MER, a detection method tailored to the specific characteristics of the target signal is expected to enhance both precision and efficiency. Accordingly, the CCW method was introduced as an experimental method built upon the shape of seismic signals generated by elephant footfalls.

\begin{figure}[!t]
 \centering
 \includegraphics[width=0.45\textwidth]{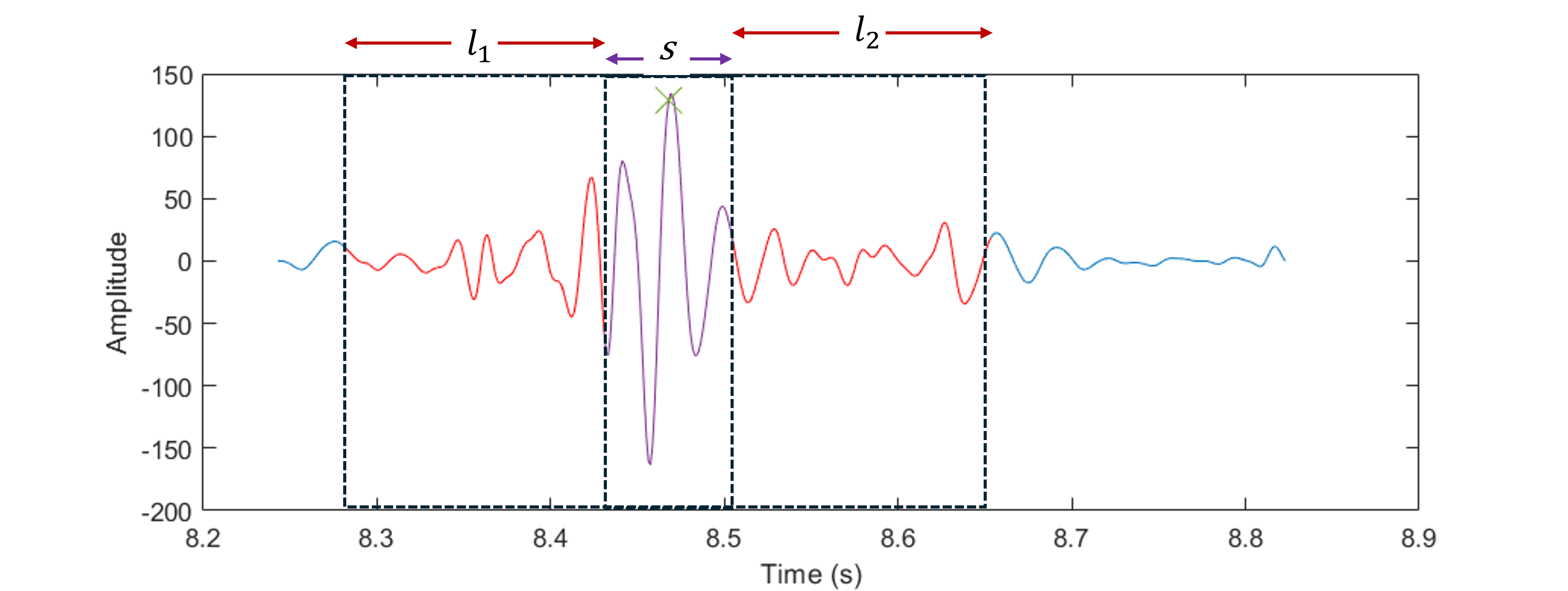}
 \caption{\hl{\textbf{Window structure of CCW method} }}
 \label{fig:CCWWindowStructure}
\end{figure}

As observed in our previous work \cite{wijayaraja2024towards} and illustrated in Fig. \ref{fig:FootfallFormPreviousStudy}, a seismic signal produced by a single elephant footstep typically exhibits a higher amplitude concentrated at the center of the waveform. This feature is further corroborated by patterns reported in prior literature \cite{wood2005using}. Based on this observation, \hl{in this study, a customized sliding window structure was designed,} consisting of a central short window flanked by two longer windows, as shown in Fig. \ref{fig:CCWWindowStructure}.

In the single elephant footstep signal depicted in Fig. \ref{fig:CCWWindowStructure}, the two flanking long windows, marked in red, have a length \hl{denoted by $l_1$ and $l_2$}, while the central short window, marked in purple, has a length denoted by $s$. Each window traverses the signal, and the ratio of average energy, $er_{ccw}[n]$, at index $n$ is \hl{computed as introduced herein in (\mbox{\ref{eq:CCW}}).} The signal function is represented by $x$.

\begin{equation}
\label{eq:CCW}
er_{ccw}[n] = \frac{\frac{1}{s} \sum_{n_1=n-\frac{s}{2}}^{n+\frac{s}{2}} \left( x[n_1] \right)^2}{\frac{1}{l_1} \sum_{n_2=n-l_1-\frac{s}{2}}^{n-\frac{s}{2}} \left( x[n_2] \right)^2 + \frac{1}{l_2} \sum_{n_3=n+\frac{s}{2}}^{n+\frac{s}{2}+l_2} \left( x[n_3] \right)^2}
\end{equation}

As the sliding window aligns with the signal pattern of the elephant footstep as depicted in Fig. \ref{fig:CCWWindowStructure}, a high, $er_{ccw}[n]$ is expected. In contrast, $er_{ccw}[n]$ remains relatively low for other signal patterns or continuous noise. Thus, potential seismic events corresponding to elephants can be effectively identified. \hl{Given the approximately equal lengths of $l_1$ and $l_2$ observed in Fig.\mbox{\ref{fig:CCWWindowStructure}}, the subsequent application of the CCW method in this study employs the simplification $l_1=l_2=l$ }

\hl{In the execution of the CCW method, the computation of the CCW energy ratio $er_{ccw}$, defined in  (\mbox{\ref{eq:CCW}}), is performed as detailed in Algorithm \mbox{\ref{alg:CCWAlgorithm}}. In this algorithm, a segment of the seismic signal with a sample length $N$ is used; this length can be adjusted according to resource constraints but should exceed the duration of a single elephant footstep. The lengths of the analysis windows ($s$ and $l$) are defined in terms of the number of discrete sampling points. The algorithm outputs the CCW energy ratio array ($er_{ccw}$).  A similar moving window technique is employed for the STA/LTA and MER methods, with the respective ratio computations adapted according to  (\mbox{\ref{eq:stalta}})  and (\mbox{\ref{eq:modified_energy_ratio}}).}

\begin{algorithm}[!t]
\caption{Optimized Computation of CCW Energy Ratio for Embedded Systems}
\label{alg:CCWAlgorithm}
\begin{algorithmic}[1]

\REQUIRE $X[1 \dots N]$: input seismic signal of length $N$  
\REQUIRE $s$: short central window length  
\REQUIRE $l$: flanking window length (for both sides)  
\ENSURE $er\_ccw[\,]$: CCW energy ratio array

\STATE $er\_ccw[1 \dots N] \leftarrow$ array of zeros
\STATE $s_{half} \leftarrow \text{floor}(s / 2)$
\STATE $n_{start} \leftarrow l + s_{half} + 1$
\STATE $n_{end} \leftarrow N - s_{half} - l$
\STATE $E_{s},E_{l1},E_{l2} $ : Variables to store average energy values
\FOR{$n = n_{start}$ to $n_{end}$}

    \STATE $E_s \leftarrow 0$
    \STATE $E_{l1} \leftarrow 0$
    \STATE $E_{l2} \leftarrow 0$
    
    \FOR{$i = n - s_{half}$ to $n + s_{half}$}
        \STATE $E_s \leftarrow E_s + X[i]^2$
    \ENDFOR
    \STATE $E_s \leftarrow E_s / s$

    \FOR{$i = n - s_{half} - l$ to $n - s_{half} - 1$}
        \STATE $E_{l1} \leftarrow E_{l1} + X[i]^2$
    \ENDFOR
    \STATE $E_{l1} \leftarrow E_{l1} / l$

    \FOR{$i = n + s_{half} + 1$ to $n + s_{half} + l$}
        \STATE $E_{l2} \leftarrow E_{l2} + X[i]^2$
    \ENDFOR
    \STATE $E_{l2} \leftarrow E_{l2} / l$

    \STATE $er\_ccw[n] \leftarrow E_s / (E_{l1} + E_{l2})$

\ENDFOR

\RETURN $er\_ccw[\,]$

\end{algorithmic}
\end{algorithm}

To select the most suitable event detection method, the three aforementioned methods were compared in terms of accuracy and execution time. The results are presented in Table \ref{tab:event_detection_comparison}.

\subsection{Selecting and Extracting Events} \label{lbl:selecting and extracting events}

The selection of potential seismic events, while minimizing background noise, is achieved by applying three empirically defined thresholds to the outputs $er[n]$, $mer[n]$ and $er_{ccw}[n]$ corresponding to event detection algorithms STA/LTA, MER and CCW, respectively. The threshold for each method was fine-tuned using reference video footage from field studies to ensure that the algorithm's output exceeded the threshold when a footfall event occurred.

Followed by, event detection, temporal segments with lengths outside the range defined by  $L_{\text{Elp\_Min}}$ and $L_{\text{Elp\_Max}}$ were discarded to remove anomalies. Subsequently, event extraction was performed in three steps: (i)  The midpoint index of each retained temporal segment was then designated as the point of interest ($PoI$),  (ii) A constant bias value ($L_{\text{Bias}}$) was introduced to adjust the midpoint, with $L_{\text{Bias}}$  which was optimized using video references. (iii) The event boundaries were computed using (\ref{eq:EventStartIndex}) and (\ref{eq:EventEndIndex}), corresponding to the start and end indices, respectively, with the extracted event defined by the segment between these indices.

\begin{equation}
\label{eq:EventStartIndex}
    \text{Starting index} = \text{PoI} + L_{\text{Bias}} -\frac{L_{\text{Elp}}}{2}
\end{equation}

\begin{equation}
\label{eq:EventEndIndex}
    \text{Ending index} = \text{PoI} + L_{\text{Bias}} + \frac{L_{\text{Elp}}}{2}
\end{equation}

\hl{The execution of the aforementioned process for event selection and extraction using the CCW method is detailed in Algorithm \mbox{\ref{alg:EventExtraction}}. The CCW energy ratio array ($er_{ccw}$) used in this algorithm is obtained from the preceding Algorithm \mbox{\ref{alg:CCWAlgorithm}}. The same procedure can be applied to the STA/LTA and MER methods by substituting the corresponding energy ratio arrays in place of $er_{ccw}$.}

\begin{algorithm}[!t]
\caption{Event Detection and Feature Extraction (CCW Method)}
\label{alg:EventExtraction}
\begin{algorithmic}[1]

\REQUIRE $X[1 \dots N]$: input seismic signal  
\REQUIRE $er\_ccw[1 \dots N]$: CCW energy ratio array  
\REQUIRE $TH_{ccw}$: threshold value for detection  
\REQUIRE $L_{\text{Elp\_Min}}$, $L_{\text{Elp\_Max}}$: minimum and maximum allowable event lengths (in samples)  
\REQUIRE $L_{\text{Bias}}$: bias offset for midpoint adjustment  
\REQUIRE $L_{\text{Elp}}$: final fixed-length event segment

\STATE $start\_flag \leftarrow$ \textbf{false}

\FOR{$n = 1$ \TO $N-1$}

    \IF{$er\_ccw[n] < TH_{ccw}$ \AND $er\_ccw[n+1] \geq TH_{ccw}$}
        \STATE $start\_index \leftarrow n$
        \STATE $start\_flag \leftarrow$ \textbf{true}
    \ENDIF

    \IF{$er\_ccw[n] \geq TH_{ccw}$ \AND $er\_ccw[n+1] < TH_{ccw}$ \AND $start\_flag$}
        \STATE $end\_index \leftarrow n$
        \STATE $start\_flag \leftarrow$ \textbf{false}

        \STATE $segment\_length \leftarrow end\_index - start\_index + 1$

        \IF{$L_{\text{Elp\_Min}} \leq segment\_length \leq L_{\text{Elp\_Max}}$}
            \STATE $PoI \leftarrow \text{floor}((start\_index + end\_index) / 2)$
            \STATE $PoI_{adj} \leftarrow PoI + L_{\text{Bias}}$
            \STATE $ExtractStart \leftarrow PoI_{adj} - \text{floor}(L_{\text{Elp}} / 2)$
            \STATE $ExtractEnd \leftarrow PoI_{adj} + \text{floor}(L_{\text{Elp}} / 2)$

            \STATE $Event \leftarrow X[ExtractStart : ExtractEnd]$

            \STATE // Perform feature extraction on the event
            \STATE FeatureExtraction($Event$)
        \ENDIF
    \ENDIF

\ENDFOR

\end{algorithmic}
\end{algorithm}

\subsection{Feature Extraction}

Feature extraction transforms extracted seismic event into structured representations that capture the distinctive characteristics, enabling the differentiation of seismic events. In this study, nine features were extracted from the identified events.

\hl{These features are listed in Table \mbox{\ref{tab:FeaturesSummery}}, and are categorized into four feature types, representing signal characteristics in different domains: temporal, spectral, pattern matching, and statistical. Moreover, a reference term was assigned to each feature, which will be referred to hereinafter in this study.}

\begin{table*}[!t]
\centering
\caption{Features used in the study}
\begin{tabular}{|l|l|l|}
\hline
\textbf{Feature Type} & \textbf{Feature} & \textbf{Reference Term} \\ \hline
\multirow{2}{*}{Temporal} & Event Length (in number of samples) & Event Length \\ \cline{2-3} 
                          & Number of Zero Crossings & Zero Crossings \\ \hline
Spectral & Predominant Frequency & Pred. Frequency \\ \hline
\multirow{4}{*}{Pattern Matching} & Maximum Cross-Correlation Value & Max Cross-Corr \\ \cline{2-3} 
                          & Cross-Correlation Value at 0 Lag & Cross-Corr (0) \\ \cline{2-3} 
                          & Mean Squared Error & MSE \\ \cline{2-3} 
                          & DTW Alignment Cost & DTW \\ \hline
\multirow{2}{*}{Statistical} & Skewness & Skewness \\ \cline{2-3} 
                          & Kurtosis & Kurtosis \\ \hline
\end{tabular}
\label{tab:FeaturesSummery}
\end{table*}

\subsubsection{Temporal Features}

A fundamental temporal characteristic of a signal is the length of the event. In this study, event length represents the original length (in samples) of the detected event, as identified by the event detection method, prior to event extraction.

Moreover, motivated by its successful application in related studies \cite{parihar2022variational}, the number of zero crossings is also employed as a temporal feature. The number of zero crossings ($Z$)  is computed using (\ref{eq:zero_crossings}). Here, $N$ denotes the length of the signal (in samples), $X[n]$ is the signal at sample $n$, and $ \mathbb{I} $ is the indicator function. The indicator function returns 1 if  $(X[n] \cdot X[n+1]) < 0$  is true, and 0 otherwise \hl{\mbox{\cite{toledo2020study}}}. 

\begin{equation}
\label{eq:zero_crossings}
Z = \sum_{n=1}^{N-1} \mathbb{I}\big[(X[n] \cdot X[n+1]) < 0\big]
\end{equation}

\subsubsection{Spectral Features}

A key characteristic of elephant footfalls is the low frequency, typically around 20 Hz \cite{o2000seismic}. Consequently, frequency-based features, such as the energy and correlation coefficients of power spectra, have been utilized to distinguish elephant footfalls from other signals  \cite{parihar2022variational,wood2005using}. Notably, the predominant frequency was successfully employed to detect elephants in our previous study \cite{wijayaraja2024towards}. The predominant frequency is calculated using (\ref{eq:predominant_frequency}), where $f_p$ represents the predominant frequency and $\mathcal{F}(X[n])$ is the Fourier Transform of the discrete signal $X[n]$, where $n$ denotes the sample index in the time domain \hl{\mbox{\cite{stoica2005spectral}}}.

\begin{equation}
\label{eq:predominant_frequency}
f_p = \arg \max_f \left| \mathcal{F}(X[n]) \right|^2
\end{equation}

\subsubsection{Pattern Matching Features}

The features categorized under 'pattern matching' rely on a pre-identified and verified footfall signal pattern. This signal pattern was cross-referenced with video footage to ensure its accuracy. To address the variability in amplitude caused by differences in gain settings and detection range, the amplitude of each pattern was normalized before further analysis.

Cross-correlation is a method that has been successfully utilized for signal identification in seismology \cite{tibi2017rapid}. Therefore, it was also considered for identifying seismic  events of elephant footfall. The cross-correlation at lag  $\tau$  \hl{between} the original signal $X[n]$ and the reference signal $Y[n]$ is represented in (\mbox{\ref{eq:cross_correlation}}) as $R_{xy}(\tau)$, where N represents the length (in samples) of the original signal ($X[n]$) \hl{\mbox{\cite{proakis2007digital}}}.

\begin{equation}
\label{eq:cross_correlation}
R_{xy}(\tau) = \sum_{n=0}^{N-1} X[n] \cdot Y[n+\tau] 
\end{equation}

However, for simplicity of implementation, the maximum cross-correlation value ($R_{\text{max}}$) and the cross-correlation value at lag 0 ($R_{xy}(0)$) were used as features. These are represented in (\ref{eq:max_cross_correlation}) and (\ref{eq:cross_correlation_zero}), respectively.

\begin{equation}
\label{eq:max_cross_correlation}
R_{\text{max}} = \max_{\tau} R_{xy}(\tau) 
\end{equation}

\begin{equation}
\label{eq:cross_correlation_zero} 
R_{xy}(0) = \sum_{n=0}^{N-1} X[n] \cdot Y[n] 
\end{equation}

Additionally, the extracted event was compared to the accurate signal pattern by calculating the mean squared error, which was also used as a feature.The mean squared error ($MSE$) is represented in (\ref{eq:mse}), where $X[n]$ is the signal, $Y[n]$ is the reference signal pattern, and $N$ is the length of both the original signal and the reference signal \hl{\mbox{\cite{wang2009mean}}}.

\begin{equation}
\label{eq:mse}
MSE = \frac{1}{N} \sum_{n=1}^N \left( X[n] - Y[n] \right)^2 \end{equation}

Dynamic Time Warping (DTW) is a method introduced for time-series pattern recognition  \cite{berndt1994using}, and has also been applied to seismology-related applications  \cite{kumar2022dynamic}. DTW calculates the sum of distances between the optimally aligned points of two signal patterns. Thus, DTW alignment cost measures the difference between the extracted event and the reference footfall signal pattern (See Fig. \ref{fig:FootfallFormPreviousStudy} ), making it a potential feature. \hl{The DTW alignment cost ($DTW(X,Y)$) is computed over the optimal warping path $P$ as shown in} (\mbox{\ref{eq:dtw}}), where $X[i]$ is the original signal value at index $i$ and $Y[j]$ is the reference signal value at index $j$ \hl{\mbox{\cite{senin2008dynamic}}}.

\begin{equation}
\label{eq:dtw} 
DTW(X, Y) = \min_P \sum_{(i, j) \in P} \| X[i] - Y[j] \| \end{equation}

\subsubsection{Statistical Features}

Finally, the statistical features of skewness and kurtosis were used for classification, as inspired by \cite{parihar2022variational}. Skewness measures the asymmetry of the distribution of the signal, while kurtosis quantifies the tailedness of the distribution. Equations (\ref{eq:skewness}) and (\ref{eq:kurtosis}) define skewness ($S$) and kurtosis ($K$), respectively. Here, $X[n]$ represents the recorded signal, $N$ is the number of samples, $\mu$ is the mean of the signal, and $\sigma$ is the standard deviation of the signal \hl{\mbox{\cite{mccrary2015implementing}}}.

\begin{equation}
\label{eq:skewness} 
S = \frac{\frac{1}{N} \sum_{n=1}^N (X[n] - \mu)^3}{\sigma^3} \end{equation}

\begin{equation}
\label{eq:kurtosis}
K = \frac{\frac{1}{N} \sum_{n=1}^N (X[n] - \mu)^4}{\sigma^4} \end{equation}

\subsection{Event Classification}
\label{lbl:event classification}
Two models namely, Artificial Neural Networks (ANN) and Support Vector Machines (SVM) were chosen for the classification process. SVM was selected for its transparency and lower computational complexity, while ANN was employed as a reference model due to its potential for higher performance, despite its increased complexity.

The SVM classifier aims to optimize a hyperplane that maximizes the margin between the two classes while minimizing classification errors. Given a dataset consisting of $N$ training samples, each sample is represented by a feature vector $x_i \in \mathbb{R}^d$ and a corresponding class label $y_i \in {-1,1}$. The optimization problem is defined as shown in (\mbox{\ref{eq:svm_optimization}}), where $W$ denotes the weight vector, $b$ represents the bias term, $C$ is the regularization parameter, and ${\xi}_i$ are slack variables introduced to account for misclassifications \hl{\mbox{\cite{norton2017soft,scholkopf2002learning}}}. The
constraints of the optimization problem are given in  (\mbox{\ref{eq:svm_constraints}}) \hl{\mbox{\cite{norton2017soft}}.}

\begin{equation} 
\label{eq:svm_optimization}
\min_{\mathbf{W},b,\xi} \frac{1}{2} ||\mathbf{W}||^2 + C \sum_{i=1}^{N} \xi_i
\end{equation}

\begin{equation}
\label{eq:svm_constraints} 
y_i (\mathbf{W}^T \mathbf{x}_i + b) \geq 1 - \xi_i, \quad \xi_i \geq 0, \quad i = 1, \dots, N.
\end{equation}

For the SVM, several kernel functions including linear, third-degree polynomial, sixth-degree polynomial, Radial Basis Function (RBF), and sigmoid were evaluated. \hl{The kernel function maps the original input vectors into a higher-dimensional feature space, where linear separation becomes feasible. It is formulated in (\mbox{\ref{eq:svm_kernel}}), where $K(x_i, x_j)$ denotes the kernel function, and $\phi(\cdot)$  represents a transformation that maps input vectors $x_i$ and $x_j$ into a higher-dimensional feature space\mbox{\cite{scholkopf2002learning}}}.

\begin{equation}
\label{eq:svm_kernel}
K(\mathbf{x}_i, \mathbf{x}_j) = \phi(\mathbf{x}_i)^T \phi(\mathbf{x}_j)
\end{equation}

To enhance the model's generalization capability and to prevent overfitting, k-fold cross-validation was utilized.\hl{ The cross-validation accuracy estimate is computed as the average of the individual accuracies obtained across all $k$ folds \mbox{\cite{kohavi1995study}}. Accordingly, the cross-validation accuracy $({CV_k})$ given in (\mbox{\ref{eq:cross_validation}})}, where $k$ represents the number of folds, and $Accuracy_i$ denotes the accuracy obtained in the $i^{th}$ fold. This method reduces variance and increases the reliability of the performance evaluation by averaging the results across multiple folds. In the conducted experiments, $k$ was empirically set to 10.

\begin{equation}
\label{eq:cross_validation}
\text{CV}_{k} = \frac{1}{k} \sum_{i=1}^{k} \text{Accuracy}_i
\end{equation}

The ANN used in this study utilizes  a sequential architecture tailored for binary classification, consisting of an input layer with 128 neurons, followed by three hidden layers with 64, 32, and 16 neurons, respectively. The Rectified Linear Unit (ReLU) activation functions were employed to enhance learning efficiency.The ReLU function was defined as (\mbox{\ref{eq:relu_activation}}), where $x$ represents the input to the activation function \hl{\mbox{\cite{he2018relu}}} .

\begin{equation}
\label{eq:relu_activation}
\text{ReLU}(x) = \max(0, x)
\end{equation}

The output layer contained a single neuron with a sigmoid activation function, mapping the network’s output to a probability range of $[0,1]$, making it suitable for binary classification. The sigmoid function ($\sigma(x)$) is given by (\ref{eq:sigmoid_function}), where $x$ is the weighted sum of neuron input \hl{\mbox{\cite{bishop2006pattern}}}. This function ensures that output can be interpreted as probabilities, facilitating the use of cross-entropy loss for training.

\begin{equation}
\label{eq:sigmoid_function}
\sigma(x) = \frac{1}{1 + e^{-x}}
\end{equation}

During feed-forward propagation, the weighted sum of inputs and biases at each layer was computed using (\ref{eq:feedforward}) \hl{\mbox{\cite{goodfellow2016deep}}}. Here $z^{(l)}$ represents the pre-activation value at layer $l$, $W^{(l)}$ is the weight matrix, $a^{(l-1)}$ represents the activation values from the previous layer, and $b^{(l)}$ is the bias vector. The activation function was then applied to obtain the final activation value as expressed in (\ref{eq:activation}), where $f(\cdot)$ represents the activation function \hl{\mbox{\cite{goodfellow2016deep}}}, that is ReLU for hidden layers and sigmoid for the output layer.

\begin{equation}
\label{eq:feedforward}
z^{(l)} = \mathbf{W}^{(l)} \mathbf{a}^{(l-1)} + \mathbf{b}^{(l)} 
\end{equation}

\begin{equation}
\label{eq:activation}
\mathbf{a}^{(l)} = f(z^{(l)}) 
\end{equation}

To improve training stability and generalization, batch normalization was applied to the first two hidden layers, while dropout layers were incorporated to prevent overfitting. Specifically, a dropout rate of 30\% was applied to the input and first hidden layer, while a 20\% dropout rate was used in the second hidden layer.  

The ANN model was trained using binary cross-entropy loss ($L$), \hl{which is a standard} loss function for binary classification problems. \hl{It is defined in (\mbox{\ref{eq:binary_crossentropy}}), where $y_i$ is the true class label, and $\hat{p}_i$ is the predicted probability of the positive class for $i^{\text{th}}$ sample  \mbox{\cite{ho2019real}}.}

\begin{equation} 
\label{eq:binary_crossentropy}
L = - \frac{1}{N} \sum_{i=1}^{N} \left[y_i \log(\hat{p}_i) + (1 - y_i) \log(1 - \hat{p}_i) \right] 
\end{equation}

The training was performed using backpropagation, where gradients were propagated backward through the network to update weights. The error term in each layer was computed as (\ref{eq:backpropagation}), where $\delta^{(l)}$ represents the error in layer $l$, $\frac{\partial L}{\partial a^{(l)}}$ is the derivative of the loss function ($L$) with respect to the activation, and  $f'(z^{(l)})$ is the derivative of the activation function \hl{\mbox{\cite{nielsen2015neural}}}. The gradient descent update rule of weight adjustment is given in (\ref{eq:weight_update}), where $\eta$ is the learning rate, $W^{(l)}$ represents the weight matrix in layer $l$, and $\frac{\partial L}{\partial W^{(l)}}$ is the gradient of the loss function ($L$) with respect to the weights \hl{\mbox{\cite{nielsen2015neural}}}. The bias update was given by (\ref{eq:bias_update}), where $b^{(l)}$ represents the biases in layer $l$, which were updated alongside the weights to optimize the network's performance \hl{\mbox{\cite{nielsen2015neural}}}.

\begin{equation}
\label{eq:backpropagation}
\delta^{(l)} = \frac{\partial L}{\partial \mathbf{a}^{(l)}} \odot f'(z^{(l)}) 
\end{equation}

\begin{equation}
\label{eq:weight_update}
W^{(l)} = W^{(l)} - \eta \frac{\partial L}{\partial W^{(l)}}
\end{equation}

\begin{equation}
\label{eq:bias_update}
b^{(l)} = b^{(l)} - \eta \delta^{(l)}
\end{equation}

For further optimization, the Adam optimizer was employed with a reduced learning rate of 0.0005 to ensure stable convergence. Adam optimizer adaptively adjusts the learning rate for each parameter using first and second-moment estimates. The weight update rule for Adam optimizer is given in (\ref{eq:adam_update}), where $w_t$ is the weight at iteration $t$, $\eta$ is the learning rate, $\hat{m}_t$ and $\hat{v}_t$ are the bias-corrected first and second-moment estimates of the gradients, and $\epsilon$ is a small constant \hl{\mbox{\cite{ruder2016overview}}}.

\begin{equation}
\label{eq:adam_update}
w_{t+1} = w_t - \frac{\eta}{\sqrt{ \hat{v}_t} + \epsilon} \hat{m}_t
\end{equation}

For training the neural network, 80\% of each dataset was allocated while 20\% was reserved for testing. The selected models were trained under two distinct scenarios: (i) using a dataset from a relatively controlled environment, (ii) from a more natural setting influenced by environmental factors such as wind. Following training, the models were tested under three different conditions: (i) for a controlled environment, (ii) for wild elephants in their natural habitat, and (iii) for a more complex scenario involving wild elephants in both natural and human-inhabited areas, closely resembling HEC-prone environments.

To evaluate the performance of all of these models under different scenarios, both accuracy and the F1 score were calculated. Accuracy measures the proportion of correctly classified instances relative to the total number of instances as in (\ref{eq:accuracy}). Here, $TP$, $TN$, $FN$, and $FP$ are confusion matrix components, which are defined in Table \ref{tab:confusion_matrix_definition}

\begin{equation}
\label{eq:accuracy}
\text{Accuracy} = \frac{TP + TN}{TP + TN + FP + FN}
\end{equation}

\begin{table*}[!t]
\centering
\caption{Confusion matrix metrics and their interpretations}
\begin{tabular}{|c|l|l|}
\hline
\textbf{Confusion Metric} & \textbf{Definition} & \textbf{Classification Scenario} \\ \hline
TP & True Positive & Correctly classifies an elephant signal as an elephant \\ \hline
TN & True Negative & Correctly classifies a non-elephant signal as not from an elephant \\ \hline
FP & False Positive & Misclassifies a non-elephant signal as from an elephant \\ \hline
FN & False Negative & Misclassifies an elephant signal as not from an elephant \\ \hline
\end{tabular}
\label{tab:confusion_matrix_definition}
\end{table*}

To provide a more comprehensive evaluation of the model’s performance, the F1-score was utilized. The F1-score represents the harmonic mean of precision and recall, offering a balanced measure between false positives and false negatives. Precision is defined in (\ref{eq:precision}) and recall is computed in  (\ref{eq:recall}). The F1-score is then calculated as shown in (\ref{eq:f1_score}), ensuring a robust evaluation of the classification model. 

\begin{equation}
\label{eq:precision}
\text{Precision} = \frac{TP}{TP + FP}
\end{equation}

\begin{equation}
\label{eq:recall}
\text{Recall} = \frac{TP}{TP + FN}
\end{equation}

\begin{equation}
\label{eq:f1_score}
\text{F1 Score} = \frac{ 2\times\text{Precision} \times \text{Recall}}{\text{Precision} + \text{Recall}}
\end{equation}

Finally, the ANN model was used for explainable AI analysis to provide interpretability to the classification results. To this end, SHapley Additive exPlanations (SHAP) values were calculated using (\ref{eq:shap_values}) \hl{\mbox{\cite{vstrumbelj2014explaining}}}.Here, $\phi_i$ represents the contribution of feature $i$ to the model's prediction, $S$ is a subset of features excluding $i$, and $f(s)$ is the output when using only the features in $S$. 
 
 \begin{equation}
\label{eq:shap_values}
\phi_i = \sum_{S \subseteq N \setminus {i}} \frac{|S|! (|N| - |S| -1)!}{|N|!} \left[ f(S \cup {i}) - f(S) \right]
\end{equation}
 
These values allowed for identifying the contribution of each feature to the model’s performance. This step was included to determine the most important features for elephant footfall identification. As such, in future implementations, this knowledge could be used to optimize the classification process by pruning less important features, thereby improving efficiency in resource-constrained implementations (i.e., embedded systems).

\subsection{Signal Acquisition Setup}

Acquiring seismic signals from elephant footfalls was challenging, particularly over long distances, due to their very low amplitudes and high susceptibility to noise \cite{fernando2020gaja}. To address this challenge,  this study employed the instrumentation system that was introduced in our previous paper \cite{wijayaraja2024towards}, which incorporates a novel geophone-sensor interface.

The geophone-sensor interface is specifically designed to enhance the detection of elephant footfall signals by amplifying the desired signals while suppressing ambient noise. The system comprises a high-sensitivity (85.8 V/m/s) 10 Hz vertical geophone, a microcontroller for signal formatting and buffering, and a battery backup to ensure continuous operation. Additionally, a Raspberry Pi module is utilized for real-time signal acquisition.

Based on our empirical knowledge form previous study  \cite{wijayaraja2024towards}, the systems were configured with three distinct gain settings: S0 = 35, S1 = 580, and S2 = 25,911. Data acquisition was conducted at an approximate sampling frequency of 880 Hz.

\begin{figure}[!t]
 \centering
 \includegraphics[width=0.4\textwidth]{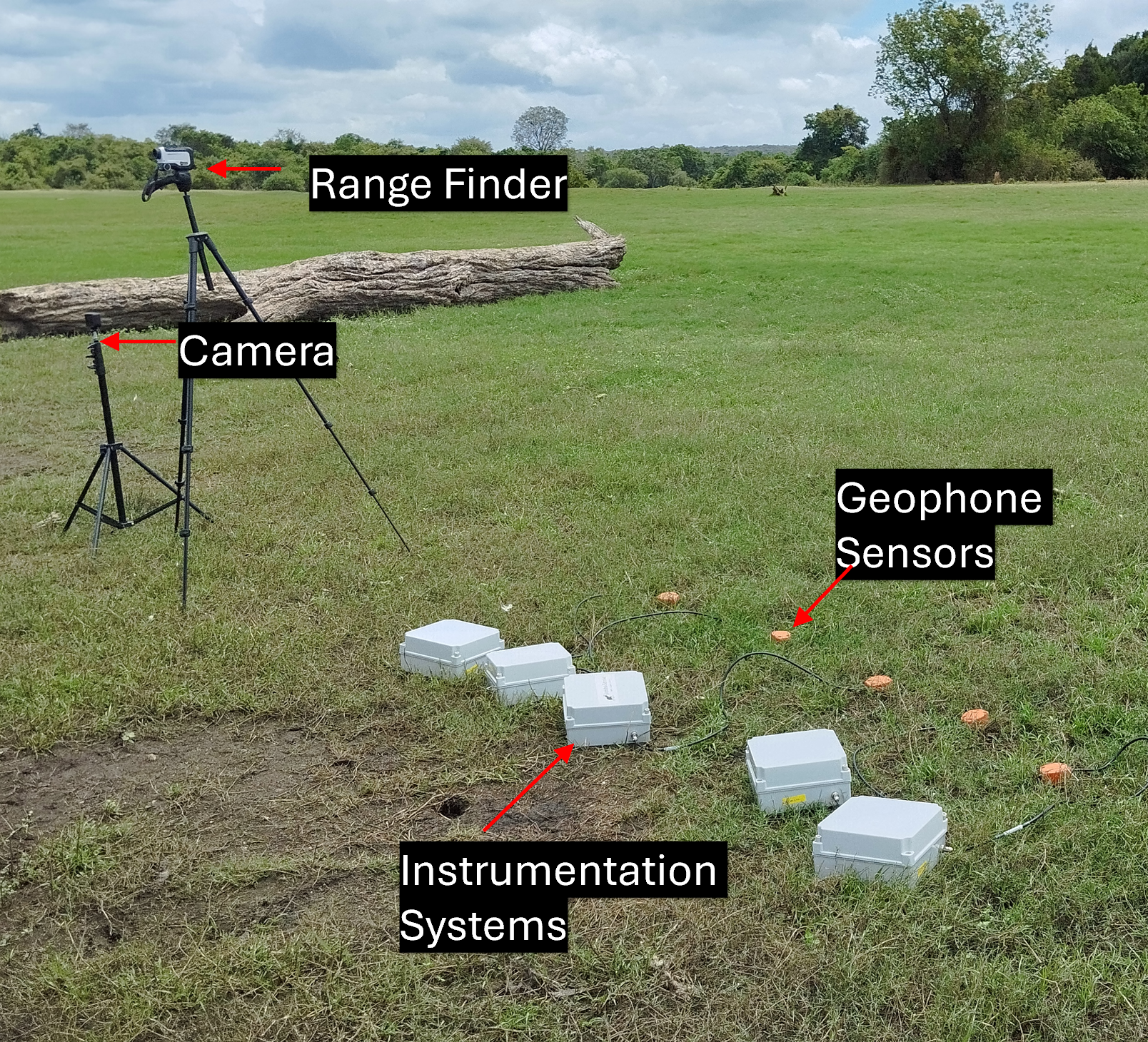}
 \caption{\textbf{Experimental setup used in field studies}}
 \label{fig:ExperimentSetup}
\end{figure}

\hl{The complete experimental setup used in the field studies is illustrated in Fig. \mbox{\ref{fig:ExperimentSetup}}. In this setup, based on the experimental scenario, multiple sensors, ranging from one to five, were deployed for seismic signal acquisition.  Furthermore,} each experiment was recorded using a camera setup to capture reference videos. Additionally, a range finder was employed to accurately measure the distances between the sensor placements and the elephants. 

\subsection{Field Studies for Data Collection}

\begin{figure}[!t]
 \centering
 \includegraphics[width=0.3\textwidth]{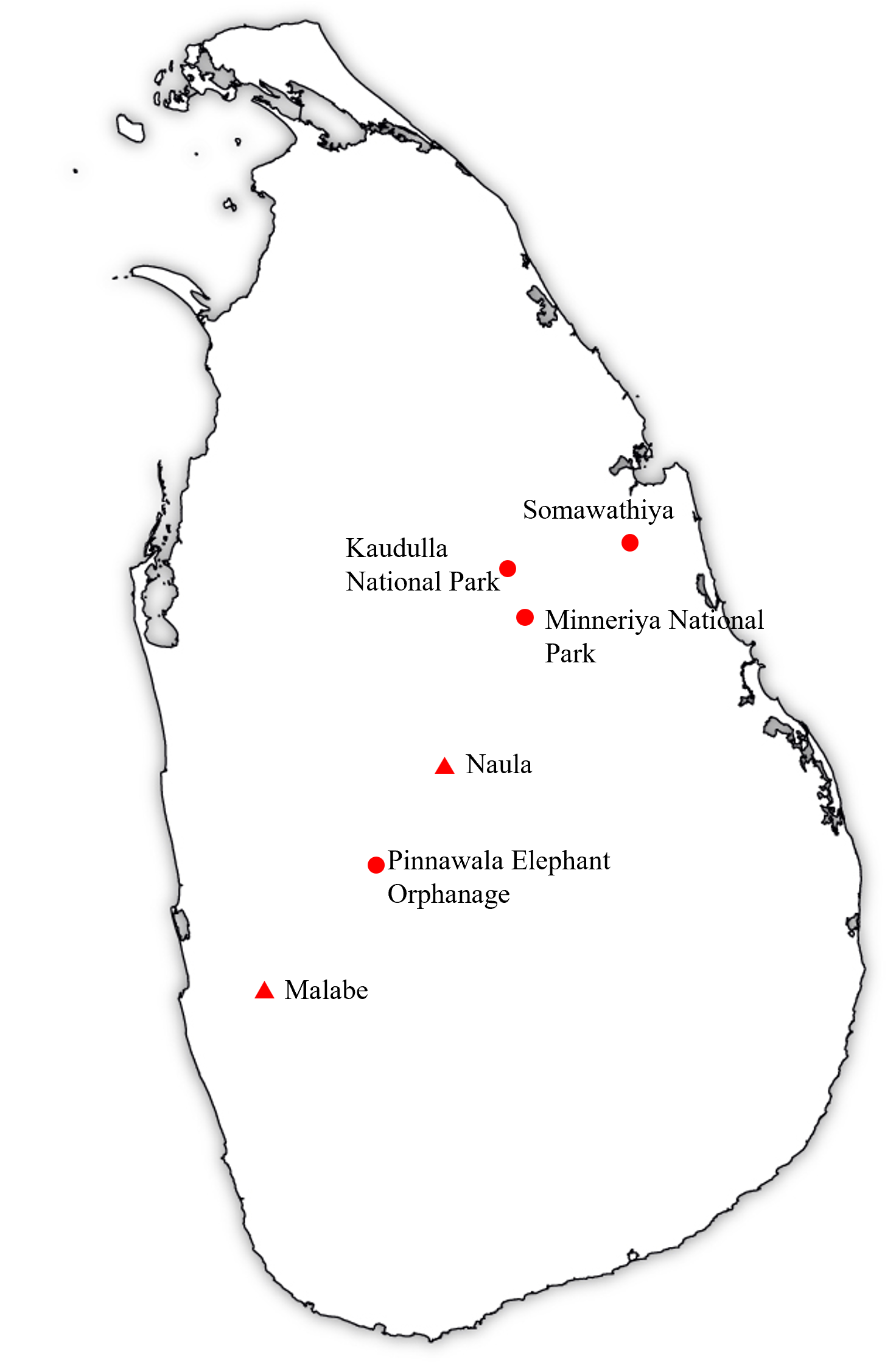}
 \caption{\textbf{Data collection sites ($\bullet$ - Sites for the collection of elephant datasets, $\blacktriangle$ - Sites for the collection of non-elephant datasets)}}
 \label{fig:Map}
\end{figure}

Using the signal acquisition setup, extensive field studies were conducted at multiple locations across Sri Lanka, as illustrated in Fig. \ref{fig:Map}. To enhance model generalization, \hl{as shown in this figure}, elephant-induced seismic signals were acquired from multiple HEC-prone regions across the country. Additionally, to improve model accuracy in real-world deployments, the non-elephant seismic dataset incorporated ambient seismic activities commonly observed in HEC-affected areas.

Field studies involving elephants were carried out under the strict supervision of the Department of National Zoological Gardens and the Department of Wildlife Conservation, Sri Lanka. All studies involving human participants were conducted with informed consent following a full explanation of the procedures. 

First, seismic signal datasets of elephant footfalls was collected under controlled conditions at the Pinnawala Elephant Orphanage, located in Rambukkana, Sri Lanka. The data acquisition involved three partially tamed elephants allowed to move freely across two designated paths: Location 1, a 100-meter stretch, and Location 2, a 160-meter path. The maximum recorded detection range was 155.6 meters at Location 2. Both sites were relatively isolated from anthropogenic activity and external seismic disturbances, enabling the collection of high-quality seismic data with minimal noise interference.  

Second, seismic signal datasets of elephant footfalls in natural habitats was collected at two sites in Sri Lanka: Kawudulla National Park (8°09'52.8"N, 80°54'47.4"E) and Minneriya National Park (8°00'57.4"N, 80°51'20.8"E). At both sites, a large group of wild elephants was observed strolling and feeding in a grassland area bordered by forest and a reservoir. In Kawudulla, seismic signals were recorded from wild elephants at distances ranging from 66.9 meters to 140 meters, while in Minneriya, the observed range was between 85.3 meters and 106 meters. 

\begin{figure}[!t]
 \centering
 \includegraphics[width=0.4\textwidth]{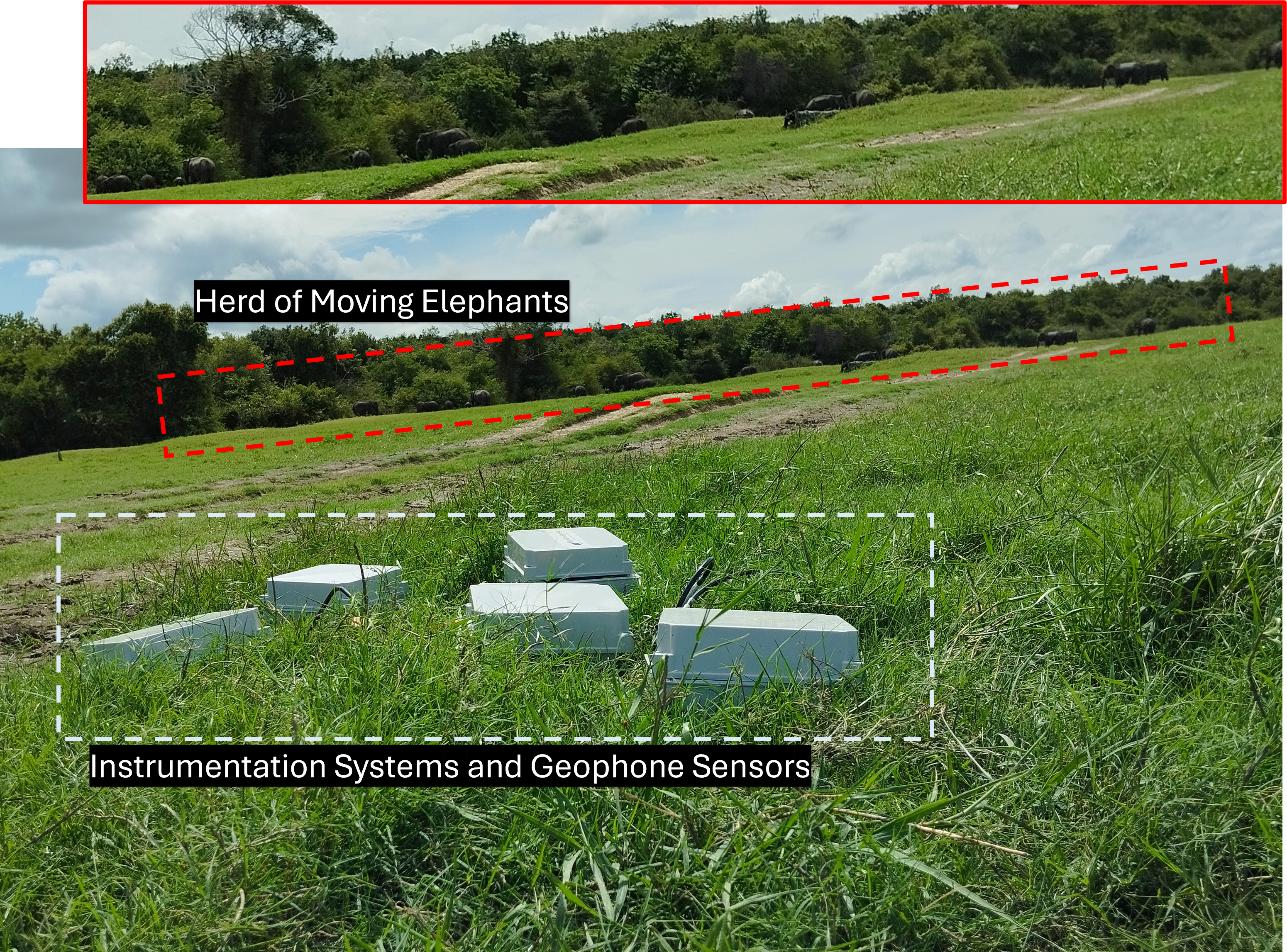}
 \caption{\textbf{Sensor and instrumentation system placement in the field study at Kawudulla}}
 \label{fig:kawudulla_sensorplacement}
\end{figure}

\hl{The positioning of the instrumentation system with the connected geophone sensors relative to the movement paths of the herd of elephants at the aforementioned long ranges is depicted in Fig. \mbox{\ref{fig:kawudulla_sensorplacement}}. Even at these challenging distances, where visual confirmation was limited due to dense vegetation, the setup enabled effective seismic signal acquisition.}

\begin{figure}[!t]
 \centering
 \includegraphics[width=0.4\textwidth]{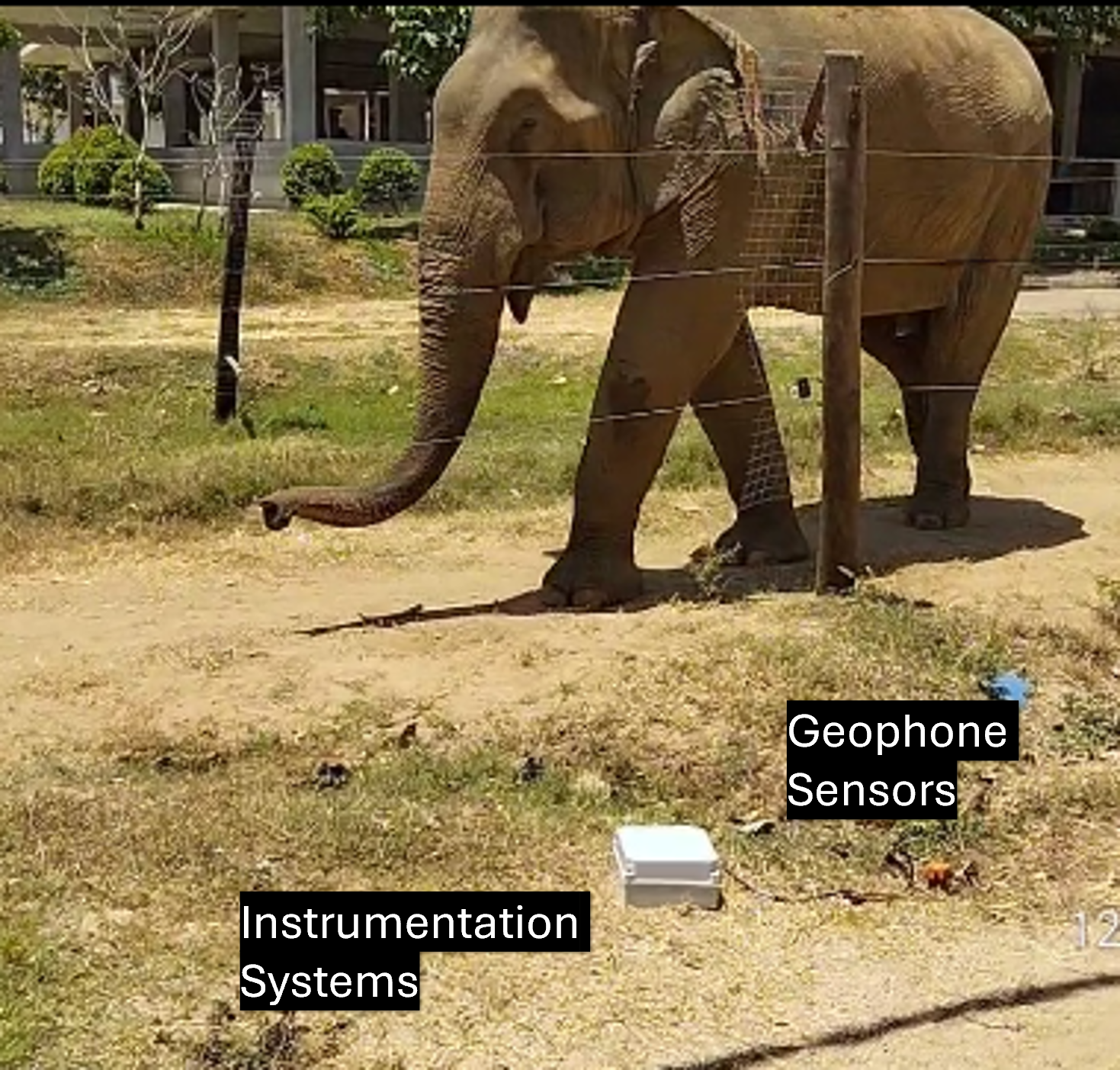}
 \caption{\textbf{Sensor and instrumentation system placement in the field study at Somawathiya)}}
 \label{fig:Somwathiya}
\end{figure}

Third, datasets of wild elephants intruding into a human-inhabited area were collected in Somawathiya, Sri Lanka, across two distinct locations within a settlement bordering a forest reserve. In Location 1, seismic signals were recorded from a single wild elephant near an electric fence, with the sensor placed in close proximity to the animal,\hl{ as shown in Fig. \mbox{\ref{fig:Somwathiya}}.} However, this dataset was affected by interference from human activities and vehicular movements. Location 2 was a relatively quiet riverbank within the same human-inhabited area, where sensors were positioned approximately 32 meters from the observed elephant. Notably, no aggressive behavior was exhibited by the elephants during either encounter.

Moreover, non-elephant datasets comprise human footfalls, recorded in the outdoor premises of the Sri Lanka Institute of Information Technology. The dataset includes three subjects, both male and female, walking along a 15-meter straight path. Seismic signals from a moving motorcycle were also recorded at the same location. 

To resemble non-elephant activity typically encountered in HEC-prone village environments, signal datasets were collected in Melpitiya, Naula, Sri Lanka (7°41'33.8"N, 80°37'39.3"E), within a paddy field adjacent to a village. In the first phase, signals were captured from the footfalls of roaming cattle. In the second phase, a combination of human and cattle footfalls, along with occasional dog footfalls, was recorded. This resulted in an unsupervised dataset that closely reflects real-world non-elephant disturbances in HEC-affected areas.

Notably, the field studies conducted in Kawudulla, Minneriya, and Naula were subjected to considerable wind conditions, while Somawathiya experienced mild wind. The intermittent movement of grass and paddy due to this wind conditions may have introduced disturbances in the recorded seismic signals.

Based on the data collected from the aforementioned studies, the datasets presented in Table \ref{tab:ExperimentSites} were prepared. The table shows the corresponding field study, the reference name for each dataset, and the gain settings used for the selected recordings.

\begin{table*}[!t]
\centering
\caption{Summary of experiment sites and corresponding datasets}
\label{tab:ExperimentSites}
\begin{tabular}{|c|c|c|c|c|}
\hline
\textbf{Experiment Site} & \textbf{Scenario} & \textbf{Subject} & \textbf{Dataset Ref.} & \textbf{Gain Settings} \\ \hline
Pinnawala & Location 1 & Elephant (Controlled) & Dt\_Pinnawala\_Loc1 & S1 \\ \hline
Pinnawala & Location 2 & Elephant (Controlled) & Dt\_Pinnawala\_Loc2 & S0, S1, S2 \\ \hline
Kawudulla & - & Elephant (Wild) & Dt\_Kawudulla & S3 \\ \hline
Minneriye & - & Elephant (Wild) & Dt\_Minneriya & S3 \\ \hline
Somawathiya & Location 1 & Elephant (Wild) & Dt\_Somawathiya\_Loc1 & S3 \\ \hline
Somawathiya & Location 2 & Elephant (Wild) & Dt\_Somawathiya\_Loc2 & S3 \\ \hline
Malabe & - & Human & Dt\_Human & S3 \\ \hline
Malabe & - & Motorcycle & Dt\_Vehicle & S3 \\ \hline
Naula & Phase 1 & Cattle & Dt\_Cattle & S3 \\ \hline
Naula & Phase 2 & Combination & Dt\_Comb & S3 \\ \hline
\end{tabular}
\end{table*}

\section{Results and Discussion}
\label{sec:Results}

\hl{ The analysis and pre-processing of seismic signals, along with the aforementioned methodologies for event detection and feature extraction, were conducted using MATLAB R2018a on a computer equipped with an Intel Core i5-1035G1 processor (1.00 GHz) and 8 GB of RAM. Subsequent classification and explainable AI (XAI) models were implemented in Python 3 on Google Colab within a CPU-only environment. The following subsections present the results corresponding to each methodological component.}

\subsection{Impact of Ambient Noise}

Across all field studies, the instrumentation system successfully recorded seismic signal patterns under the given gain settings. To attenuate high-frequency noise while minimizing distortion, a 6th-order low-pass Butterworth filter with a cutoff frequency of 80 Hz was applied initially.

\begin{figure}[!t]
 \centering
\includegraphics[width=0.4\textwidth]{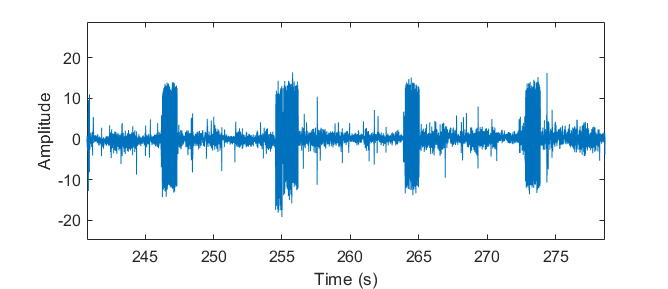}
 \caption{\textbf{Time-domain representation of noise observed in the field study at Kawudulla.}}
 \label{fig:windnoise}
\end{figure}

\begin{figure}[!t]
 \centering
 \includegraphics[width=0.4\textwidth]{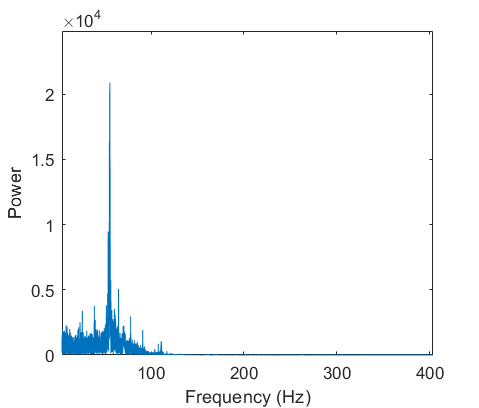}
 \caption{\textbf{Power spectrum representing the noise observed in the field study at Kawudulla.}}
 \label{fig:windnoisespectrum}
\end{figure}

However, as shown in Fig.\ref{fig:windnoise} (corresponding power spectrum in Fig.\ref{fig:windnoisespectrum}), it can be observed that data gathered from Kawudulla, Minneriya, and Naula exhibit additional noise, characterized by an abnormal peak at approximately 55 Hz. A similar noise artefact was also detected in the Somawathiya field study, albeit at a lower amplitude. With the support of the video references, it was identified that the noise was caused by the periodic movement of vegetation induced by wind.

To mitigate the identified noise, a band-stop filter was applied with stopband frequencies at 50 Hz and 60 Hz. The filter’s order and cutoff frequencies were designed using Butterworth parameters to achieve a 60 dB attenuation, effectively suppressing the targeted noise. 
The recovered signal from Kawudulla, Minneriya, and Somawathiya then provided the expected pattern for elephant footfalls, as shown in Fig. \ref{fig:BestSignalPattern}.

\subsection{Footfall Pattern of Elephants}

\begin{figure}[!t]
 \centering
 \includegraphics[width=0.45\textwidth]{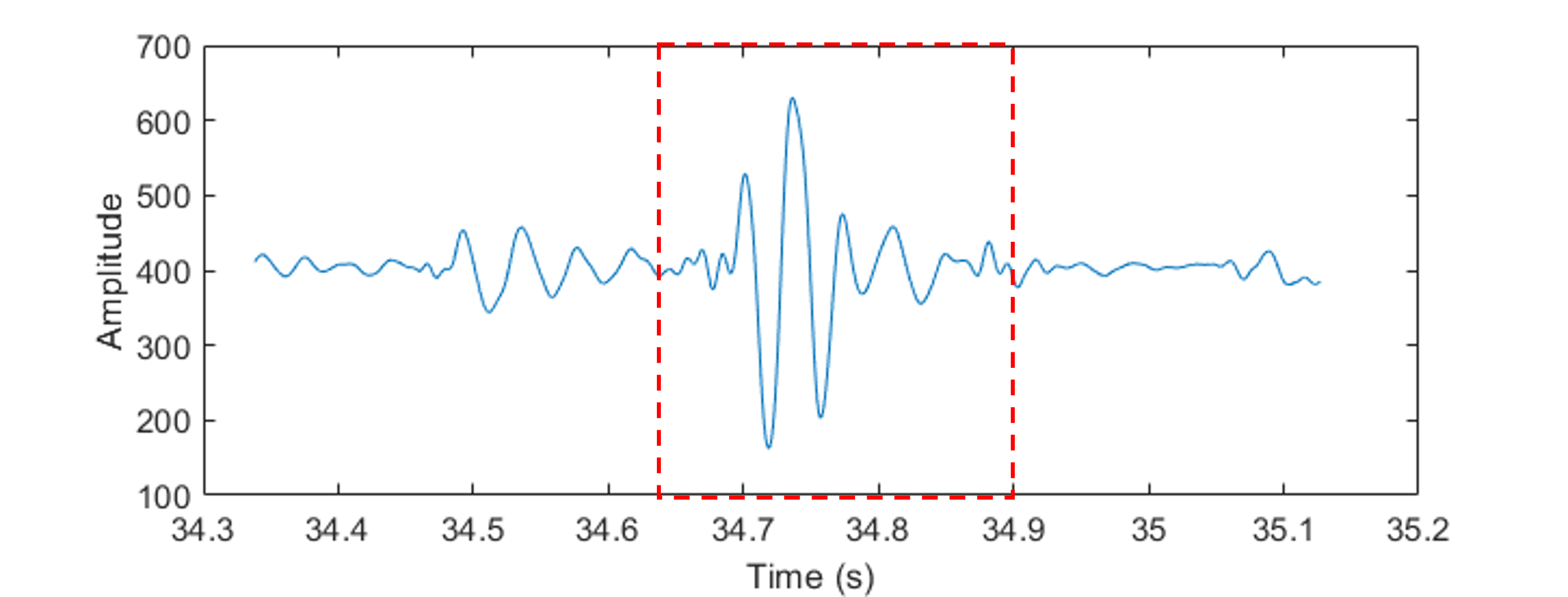}
 \caption{\textbf{Seismic signal pattern of elephant footfall}}
 \label{fig:BestSignalPattern}
\end{figure}

\begin{figure}[!t]
 \centering
 \includegraphics[width=0.45\textwidth]{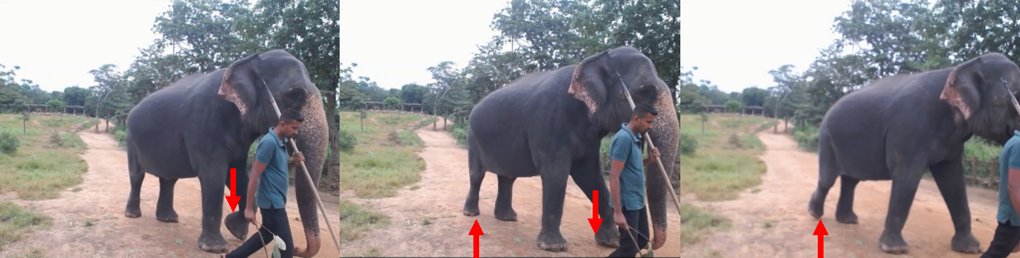}
 \caption{\textbf{Footfall of elephants}}
 \label{fig:ElephantFootfallVedio}
\end{figure}

In relatively controlled settings, where ambient noise is less, the repetitive footfall signal pattern of an elephant is depicted in Fig. \ref{fig:BestSignalPattern}. A closer inspection of this pattern reveals that the signal contains two distinct patterns with varying amplitudes, likely resulting from the movement of both the front and rear legs (see Fig. \ref{fig:ElephantFootfallVedio}). This hypothesis aligns with findings in \cite{wood2005using}, which confirms that the footfall from the front leg generates a larger amplitude compared to the rear leg. However, in a noisy environment, only the segment corresponding to the front leg could be frequently observed (the segment highlighted in Fig. \ref{fig:BestSignalPattern}). Therefore, this segment was considered as the characteristic footfall signature of elephants for the remainder of the study.

After manually inspecting the events, the average length of the considered footfall signature of elephants was found to be approximately 190 data samples. As the sampling frequency is 880 Hz for the setup, the average temporal length of the signal $L_{Elp}$ is calculated as 215.90 ms. This aligns with previous studies such as \cite{o2000seismic}, which has reported that the typical elephant footfall length ranges from 103 ms to 250 ms. Based on the observations in this study, the minimum and maximum footfall durations denoted as $L_{Elp\_Min}$ and $L_{Elp\_Max}$   are considered to be 75.00 ms and 354.55 ms, respectively. These parameters were utilized in event detection.

\subsection{Evaluation of Event Detection Methods}

To evaluate the performance of event detection methods, a 5.2-second temporal segment of the signal recording was selected, and the corresponding video footage within this time frame was analyzed. Within this segment, ten footsteps were visually identified. These events were cross-referenced and confirmed with the identifiable footsteps in the video footage. The confirmed footsteps within this segment were used as the ground truth.

\begin{figure}[!t]
 \centering
 \includegraphics[width=0.45\textwidth]{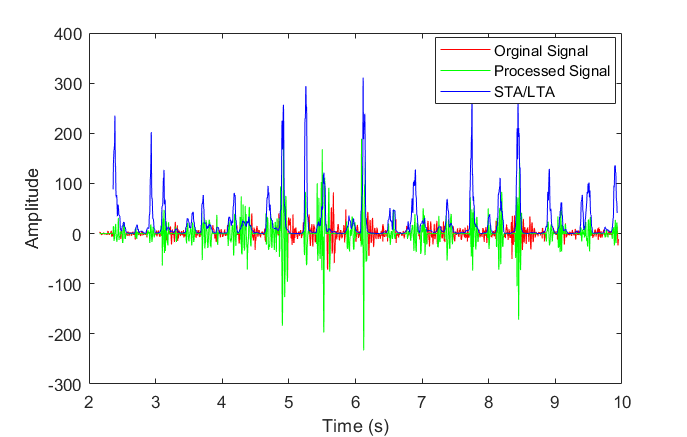}
 \caption{\textbf{Results of STA/LTA method}}
 \label{fig:STALTAResults}
\end{figure}

\begin{figure}[!t]
 \centering
 \includegraphics[width=0.45\textwidth]{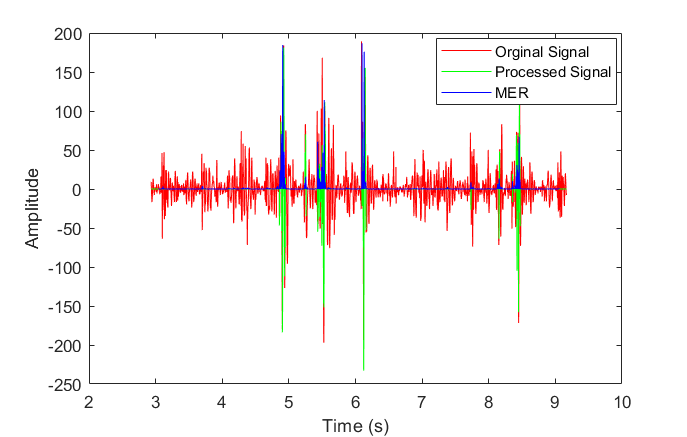}
 \caption{\textbf{Results of MER method}}
 \label{fig:MERResults}
\end{figure}

\begin{figure}[!t]
 \centering
 \includegraphics[width=0.45\textwidth]{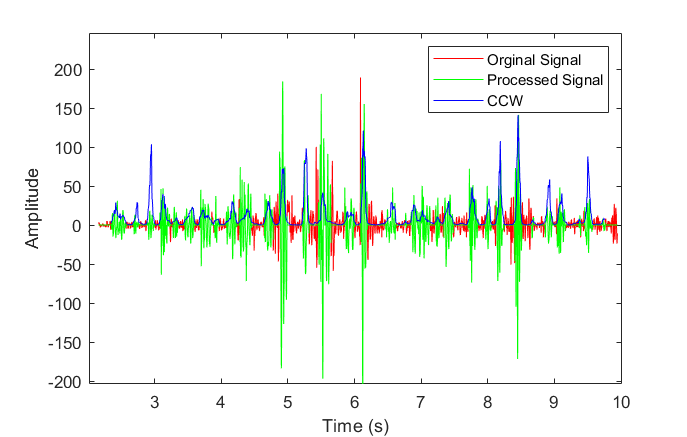}
 \caption{\textbf{Results of CCW method}}
 \label{fig:CCWesults}
\end{figure}

Subsequently, this segment was processed using three event detection methods: STA/LTA, MER, and CCW. The results from each event detection method were obtained, as illustrated in Fig. \ref{fig:STALTAResults}, Fig. \ref{fig:MERResults}, and Fig. \ref{fig:CCWesults}, which present the corresponding outputs. In these figures, the time span from 3.4 s to 8.6 s represents the selected temporal segment. A comparison of the results is provided in Table \ref{tab:event_detection_comparison}.

\begin{table}[!t]
    \centering
    \caption{Comparison of event detection methods based on detection accuracy and execution time}
    \label{tab:event_detection_comparison}
        \begin{tabular}{|c|c|c|c|c|}
        \hline
        \textbf{Method} & \textbf{Detected} & \textbf{Missed} & \textbf{Merged} & \textbf{Execution Time (ms)} \\ \hline
        STA/LTA & 10 & 0 & 2 & 13.84 \\ \hline
        MER & 4 & 6 & 0 & 14.42 \\ \hline
        CCW & 10 & 0 & 1 & 16.48 \\ \hline
        \end{tabular}
\end{table}

\hl{The detection performance of each event detection method for the selected time span is represented in Table \mbox{\ref{tab:event_detection_comparison}}. Accordingly,} both the STA/LTA and CCW methods successfully detected all the identified footsteps. However, the MER method was unsuccessful, missing six events. During the detection process, there were instances where the signal was captured but merged with additional seismic components. These events were recorded under the column titled 'Merged'. A merged event can introduce deviations in the $PoI$ during event extraction, leading to inaccurate results in subsequent processing. Within the limited time frame, STA/LTA detects two instances, while CCW identifies one, leading to the hypothesis that CCW is superior in event separation. However, it must be noted that even minor irregular foot movements by elephants can generate unpredictable seismic patterns. As a result, the authors are unable to confirm false detections in these scenarios, and false detections were not considered as factors for evaluating performance.

The average execution time of each method was included as a criterion for evaluating performance (see Table \ref{tab:event_detection_comparison}). Among the three methods, the STA/LTA method demonstrated the lowest execution time. Although the CCW method showed acceptable results, an assessment of both detection accuracy and execution time confirmed that the STA/LTA method exhibited the desirable performance in the given context of constrained computing resources. Consequently, it was selected for further processing.

Using the STA/LTA method, potential events were identified and extracted from all the datasets listed in Table \ref{tab:ExperimentSites}. As outlined in the selecting and extracting events (refer subsection \ref{lbl:selecting and extracting events}),  events with a temporal length shorter than, $L_{Elp\_Min}$ (75.00 ms) or exceeding $L_{Elp\_Max}$ (354.55 ms) were rejected.

\subsection{Results of Classification}

Following a comprehensive examination of events extracted from the field study in Pinnawala, specifically from the datasets \texttt{Dt\_Pinnawala\_Loc1} and \texttt{Dt\_Pinnawala\_Loc2} (refer Table \ref{tab:ExperimentSites}), each elephant footstep was manually annotated with video references to ensure accuracy. However, in field studies conducted in Kawudulla, Minneriya and Somawathiya, precise identification of each individual footstep was not achievable due to obstacles that hindered close observation in the natural environment. Instead, seismic signals from one-minute time intervals where footfalls were visible in the video footage were annotated. Furthermore, these annotations were validated through manual verification and cross-referenced with the pattern illustrated in Fig. \ref{fig:ElephantFootfallVedio}. All extracted events from field studies conducted in Malabe and Naula (events from non-elephant seismic datasets) were subsequently annotated as 'non-elephant events'.

The number of samples (extracted seismic events) of each dataset is presented in Table \ref{tab:dataset_samples_count} \footnote{For these experiments, only precisely matched elephant footfalls were selected from identified elephant footfall events}. It was observed that the \texttt{Dt\_Cattle} dataset exhibited a substantially higher number of recorded events, posing a potential risk of class imbalance during model training. To address this, random under-sampling was applied to this dataset to align with the mean event count of the other datasets. 

\begin{table}[!t]
    \centering
    \caption{Number of samples corresponding to each dataset}
    \label{tab:dataset_samples_count}
    \begin{tabular}{|c|c|}
        \hline
        \textbf{Dataset Reference} & \textbf{Number of Samples} \\ \hline
        Dt\_Pinnawala\_Loc1   & 96  \\ \hline
        Dt\_Pinnawala\_Loc2   & 132 \\ \hline
        Dt\_Kawudulla        & 41  \\ \hline
        Dt\_Minneriya        & 14  \\ \hline
        Dt\_Somawathiya\_Loc1 & 23  \\ \hline
        Dt\_Somawathiya\_Loc2 & 9   \\ \hline
        Dt\_Human            & 92  \\ \hline
        Dt\_Vehicle          & 7   \\ \hline
        Dt\_Cattle           & 750 \\ \hline
        Dt\_Comb             & 767 \\ \hline
    \end{tabular}
\end{table}
\begin{table}[!t]
    \centering
    \caption{Summary of training and testing cases with corresponding datasets}
    \begin{tabular}{|l|l|l|}
        \hline
        \textbf{Case Name}  & \textbf{Class}    & \textbf{Dataset Used}            \\ \hline
        \multirow{4}{*}{Train\_Case\_1} & Positive   & Dt\_Pinnawala\_Loc2            \\ \cline{2-3}
                                        & \multirow{3}{*}{Negative} & Dt\_Human            \\  
                                        &                           & Dt\_Cattle           \\  
                                        &                           & Dt\_Vehicle          \\ \hline
        \multirow{3}{*}{Train\_Case\_2} & Positive   & Dt\_Pinnawala\_Loc2            \\ \cline{2-3}
                                        & \multirow{2}{*}{Negative} & Dt\_Human            \\  
                                        &                           & Dt\_Vehicle          \\ \hline
        \multirow{2}{*}{Test\_Case\_1}  & Positive   & Dt\_Pinnawala\_Loc1            \\ \cline{2-3}
                                        & Negative   & Dt\_Comb                        \\ \hline
        \multirow{2}{*}{Test\_Case\_2}  & Positive   & Dt\_Kawudulla                  \\ \cline{2-3}
                                        & Negative   & Dt\_Comb                        \\ \hline
        \multirow{5}{*}{Test\_Case\_3}  & \multirow{4}{*}{Positive} & Dt\_Kawudulla       \\  
                                        &                           & Dt\_Minneriya       \\  
                                        &                           & Dt\_Somawathiya\_Loc1 \\  
                                        &                           & Dt\_Somawathiya\_Loc2 \\ \cline{2-3}
                                        & Negative   & Dt\_Comb                        \\ \hline
    \end{tabular}
    \label{tab:CasesForClassification}
\end{table}
\begin{table*}[h!]
    \centering
    \caption{Classification results for STA/LTA as an event detection method}
    \resizebox{\textwidth}{!}{ 
    \begin{tabular}{|c|l|c|c|c|c|c|c|c|c|c|}
        \hline
        \multirow{2}{*}{\textbf{Model ID}} & \multirow{2}{*}{\textbf{Classifier Used}} & \multirow{2}{*}{\textbf{Training Case}} & \multicolumn{2}{c|}{\textbf{Test Dataset}} & \multicolumn{2}{c|}{\textbf{Test Case 1}} & \multicolumn{2}{c|}{\textbf{Test Case 2}} & \multicolumn{2}{c|}{\textbf{Test Case 3}} \\ \cline{4-11} 
                                          &                                     &                                           & \textbf{Acc.} & \textbf{F1} & \textbf{Acc.} & \textbf{F1} & \textbf{Acc.} & \textbf{F1} & \textbf{Acc.} & \textbf{F1} \\ \hline
        A1                                & SVM (Kernel: Linear)               & \multirow{6}{*}{Train\_Case\_1}           & 89\%          & 89\%        & \textbf{81\%}& \textbf{85\%}& 69\%          & 56\%        & 67\%          & 74\%        \\ \cline{1-2} \cline{4-11} 
        A2                                & SVM (Kernel: Poly, degree=3)       &                                           & 86\%          & 86\%        & 73\%          & 81\%        & 64\%          & 27\%        & 70\%          & 79\%        \\ \cline{1-2} \cline{4-11} 
        A3                                & SVM (Kernel: Poly, degree=6)       &                                           & 77\%          & 79\%        & 72\%          & 81\%        & 62\%          & 23\%        & 64\%          & 76\%        \\ \cline{1-2} \cline{4-11} 
        A4                                & SVM (Kernel: RBF)                  &                                           & \textbf{90\%}& \textbf{90\%}& \textbf{80\%}& \textbf{85\%}& \textbf{73\%}& \textbf{60\%}
& \textbf{70\%}& \textbf{76\%}\\ \cline{1-2} \cline{4-11} 
        A5                                & SVM (Kernel: Sigmoid)              &                                           & 85\%          & 85\%        & 75\%          & 80\%        & 65\%          & 54\%        & 59\%          & 66\%        \\ \cline{1-2} \cline{4-11} 
        A6                                & ANN                                 &                                           & \textbf{100\%}& \textbf{100\%}& \textbf{70\%}& \textbf{70\%}& \textbf{66\%}& \textbf{54\%}& \textbf{64\%}& \textbf{61\%}\\ \hline
        B1                                & SVM (Kernel: Linear)               & \multirow{6}{*}{Train\_Case\_2}           & 99\%          & 100\%       & \textbf{81\%}& 85\%        & 69\%          & 56\%        & 67\%          & 74\%        \\ \cline{1-2} \cline{4-11} 
        B2                                & SVM (Kernel: Poly, degree=3)       &                                           & 100\%         & 100\%       & 73\%          & 81\%        & 64\%          & 27\%        & 70\%          & 79\%        \\ \cline{1-2} \cline{4-11} 
        B3                                & SVM (Kernel: Poly, degree=6)       &                                           & 82\%          & 86\%        & 72\%          & 81\%        & 62\%          & 23\%        & 64\%          & 76\%        \\ \cline{1-2} \cline{4-11} 
        B4                                & SVM (Kernel: RBF)                  &                                           & \textbf{99\%}& \textbf{99\%}& \textbf{80\%}& \textbf{85\%}& \textbf{73\%}& \textbf{60\%}& \textbf{70\%}& \textbf{76\%}\\ \cline{1-2} \cline{4-11} 
        B5                                & SVM (Kernel: Sigmoid)              &                                           & \textbf{100\%}& \textbf{100\%}& 75\%          & 80\%        & 65\%          & 54\%        & 59\%          & 66\%        \\ \cline{1-2} \cline{4-11} 
        B6                                & ANN                                 &                                           & \textbf{100\%}& \textbf{100\%}& \textbf{78\%}& \textbf{83\%}& \textbf{56\%}& \textbf{55\%}& \textbf{60\%}& \textbf{66\%}\\ \hline
    \end{tabular}
    }
    \label{tab:classification_results}
\end{table*}
The values for the feature specified in Section \ref{tab:FeaturesSummery} were computed for each extracted event. These features were then systematically categorized as specified in Table \ref{tab:CasesForClassification}. \texttt{Train\_Case\_1} and \texttt{Train\_Case\_2} were used to train two distinct versions of each classifier. \texttt{Train\_Case\_1} included a negative dataset, \texttt{Dt\_Cattle}, containing noise disturbances such as wind and grass movement, which are common in HEC-prone natural environments in Sri Lanka. In contrast, \texttt{Train\_Case\_2} comprised a dataset from a relatively noise-free setting. \texttt{Test\_Case\_1}, \texttt{Test\_Case\_2}, and \texttt{Test\_Case\_3} were used to evaluate the trained classifiers, representing the following scenarios: (1) partially tamed elephants in controlled settings, (2) wild elephants in their natural habitats, and (3) wild elephants in both natural and human-influenced environments.

The classification was performed as outlined in the methodology section,  using both ANN and SVM (refer \ref{lbl:event classification}). The SVM was trained using five different kernels. Each model was trained on two distinct versions of the dataset. The classification results are presented in Table \ref{tab:classification_results}. 

In Table \ref{tab:classification_results}, for SVM, the 'Test Dataset' column shows performance across all test folds, \hl{obtained using k-fold cross-validation,} where each fold serves as a validation set in successive iterations. For ANN, it represents the model's performance on the 20\% subset of the dataset reserved for evaluation after training. \texttt{Test\_Case\_1}, \texttt{Test\_Case\_2}, and \texttt{Test\_Case\_3} correspond to the evaluation of the classifiers under the respective scenarios described earlier.

According to Table \ref{tab:classification_results}, the performance evaluation using the Test Dataset demonstrated that the ANN model, serving as the reference benchmark, achieved the highest performance with an accuracy and F1 score of 100 \%. This outcome aligns with expectations, given ANN’s capacity to model complex patterns and fit the training data effectively. Among the SVM models, the SVM with an RBF kernel exhibited the best performance, achieving 99 \% accuracy and F1 score when trained on \texttt{Train\_Case\_2} (\texttt{Model B4}), compared to 89 \% accuracy and F1 score for \texttt{Train\_Case\_1} (\texttt{Model A4}). This finding highlights that the controlled settings generally lead to superior performance due to reduced noise interference.

In \texttt{Test\_Case\_1}, which evaluates model performance in a controlled test setting, the ANN trained on \texttt{Train\_Case\_2} (\texttt{Model B6}) outperformed the ANN trained on \texttt{Train\_Case\_1} (\texttt{Model A6}), reinforcing the earlier observation regarding the benefits of controlled conditions. However, despite this advantage, ANN experienced a substantial drop in accuracy and F1 score compared to the Test Dataset, suggesting that its higher complexity may have led to overfitting, reducing its robustness in varying conditions.

Although SVM models also exhibited a performance degradation in \texttt{Test\_Case\_1}, the decrease was notably less severe compared to ANN. In terms of accuracy, the SVM with a Linear kernel (\texttt{Models A1} and \texttt{B1}) marginally outperformed the RBF kernel. However, when considering the F1 score, both SVM with Linear and RBF kernels achieved 85 \%, the highest among all models, confirming that the SVM with an RBF kernel maintained satisfactory performance.

When both models were applied to elephant data in natural environments in \texttt{Test\_Case2}, the ANN model trained on \texttt{Train\_Case\_1} (\texttt{Model A6}) outperformed the model trained with \texttt{Train\_Case\_2} (\texttt{Model B6}), which was the opposite for the relatively controlled settings.  This outcome suggests that the inclusion of the \texttt{Dt\_Cattle} dataset enhanced the model performance. It is hypothesized that natural phenomena, such as wind or resulting grass movement in corresponding field studies contributed to improving the model's generalization capability.

In contrast, the performance of the SVM with the RBF kernel remained consistent regardless of the training dataset; indicating that SVM primarily relies on support vectors to define decision boundaries, making it less sensitive to dataset variations and more reliant on fundamental feature separability. Notably, the SVM model with the RBF kernel yielded the best results in the natural environment with an accuracy of 73\% and F1 score of 60\%.
\begin{table*}[!t]
    \centering
    \caption{Classification Performance Using CCW and STA/LTA Event Detection Methods}
    \label{tab:CCW_Comparison}
    \begin{tabular}{|c|c|c|c|c|c|}
        \hline
        \textbf{Classifier} & \textbf{Training Case} & \multicolumn{2}{c|}{\textbf{CCW}} & \multicolumn{2}{c|}{\textbf{STA/LTA}} \\  
        \cline{3-6}
        &  & \textbf{Accuracy} & \textbf{F1 Score} & \textbf{Accuracy} & \textbf{F1 Score} \\  
        \hline
        \multirow{2}{*}{\textbf{SVM (RBF Kernel)}} & Train Case 1 & 88\% & 87\% & 90\% & 90\% \\  
        \cline{2-6} 
        & Train Case 2 & 99\% & 99\% & 99\% & 99\% \\  
        \hline
        \multirow{2}{*}{\textbf{ANN}} & Train Case 1 & 92\% & 92\% & 100\% & 100\% \\  
        \cline{2-6} 
        & Train Case 2 & 100\% & 100\% & 100\% & 100\% \\  
        \hline
    \end{tabular}
\end{table*}

\hl{In comparison with \mbox{\cite{parihar2025hybrid}} (mentioned in Table \mbox{\ref{tab:classification_lit_summery}}), a similar experiment conducted in natural environments for Asian elephants using dedicated embedded systems, this study was able to detect elephants from over 140 m, whereas \mbox{\cite{parihar2025hybrid}} was limited to short detection ranges such as 50 m\footnote{The exact detection range was not specified; however, sensors were placed ~50 meters from electrified railway lines, the expected crossing points of elephants}. Thereby surpassing existing range benchmarks for dedicated embedded systems.}

\hl{Moreover, when comparing this study to a study that employed high-end seismometers \mbox{\cite{steinmann2025decoding}} and reported a comparable accuracy of 77\% at a range of 150 m, this work, even using relatively low-cost, custom embedded systems, achieved a comparable result with only a 4\% reduction in accuracy at a similar range of 140 meters.} These results therefore demonstrate a competitive balance between accuracy, detection range, and cost-effectiveness, supporting the authors' motivation to implement this approach as a viable solution for the HEC problem in natural environments. \hl{The results indicate competitive performance relative to prior studies, summarized in Table \mbox{\ref{tab:classification_lit_summery}}.}

In \texttt{Test\_Case\_3}, the most challenging test case, ANN which was trained on \texttt{Train\_Case\_1} (\texttt{Model A6}) (accounted for natural phenomena), achieved 64\% accuracy and an F1 score of 76\%. However, despite the difficult conditions, the SVM with the RBF kernel (\texttt{Models A4} and \texttt{B4}) outperformed ANN, achieving 70\% accuracy and an F1 score of 76\%, the highest among all models for this test case. This result underscores the robustness of the SVM model in complex environments. Most importantly, this is the first study to experiment with seismic signal classification on wild Sri Lankan elephants that entered human-inhabited areas, closely resembling real-world HEC scenarios.

For further analysis, the execution times of the SVM with the RBF kernel and ANN were compared across three test cases and both training sets. The ANN model exhibited an average execution time of 230.7 ms, whereas the SVM with the RBF kernel required only 147.69 ms. Overall, the SVM with the RBF kernel achieved high classification performance while maintaining a significantly lower execution time, reinforcing its suitability for deployment in resource-constrained implementations.

In summary, the ANN model performed well under its trained conditions but struggled to generalize, especially for natural environments. The SVM with the RBF kernel demonstrated stable performance across all test scenarios, including those closely resembling HEC-prone areas. Consequently, while ANN excels under ideal training conditions, SVM with the RBF kernel proves to be more reliable for practical deployment in a real-world environment.

\subsection{Performance and Feasibility of CCW}

CCW demonstrated comparatively accurate event detection (refer Table \ref{tab:event_detection_comparison}), prompting further evaluation of its classification performance as an event detection method. The classification outcomes using CCW were comparable to those obtained with the STA/LTA method, as shown in Table \ref{tab:CCW_Comparison}. The comparison includes the best-performing classifier: SVM with RBF kernel, and the reference model: ANN. Each model was trained and tested using \texttt{Train\_Case\_1} and \texttt{Train\_Case\_2} with both CCW and STA/LTA as event detection methods.

The comparative analysis in Table \ref{tab:CCW_Comparison} indicates that both event detection methods yield nearly identical classification performance, particularly under \texttt{Train\_Case\_2}. The CCW method demonstrates exceptionally high accuracy and F1 scores, achieving 99\% for SVM with an RBF kernel and 100\% for ANN, aligning with the results obtained using STA/LTA. These findings confirm that CCW is as effective as STA/LTA in relatively controlled settings.

However, under the more challenging \texttt{Train\_Case\_1}, a slight decline in performance is observed with the CCW method relative  to STA/LTA. While the ANN model exhibits an 8\% reduction in both accuracy and F1 score, the performance drop in the more generalized SVM model is relatively minor, with accuracy decreasing by only 2\% and F1 score by 3\%. This variation is expected in complex natural scenarios leading to a conclusion that environmental factors may introduce additional uncertainties.

The slight reduction in performance observed under \texttt{Train\_Case\_1} is likely to be caused by the fact that the \texttt{Dt\_Cattle} dataset in \texttt{Train\_Case\_1} contains non-elephant events with structural similarities to the pattern \footnote{A similar anomaly is further investigated under feature impact assessment.} in Fig. \ref{fig:CCWWindowStructure}. As the CCW method is still in its early stages, further investigations are necessary to confirm these assumptions and to enhance its robustness across diverse datasets which will be a significant future work of this study.

However, it is worth mentioning that even in its early stages, the CCW method demonstrated competitive classification performance compared to other event detection approaches. Despite its increased computational complexity, preliminary experimental results indicate that CCW outperformed existing methods in specific aspects, particularly in its resilience to event merging (refer to Table \ref{tab:CCW_Comparison}). These findings highlight the strong potential of CCW for real-world applications. Further optimization of the algorithm and fine-tuning of its parameters could enhance classification accuracy and improve its feasibility for deployment in embedded systems.

\subsection{Feature Impact Assessment}

\begin{figure}[!t]
 \centering \includegraphics[width=0.45\textwidth]{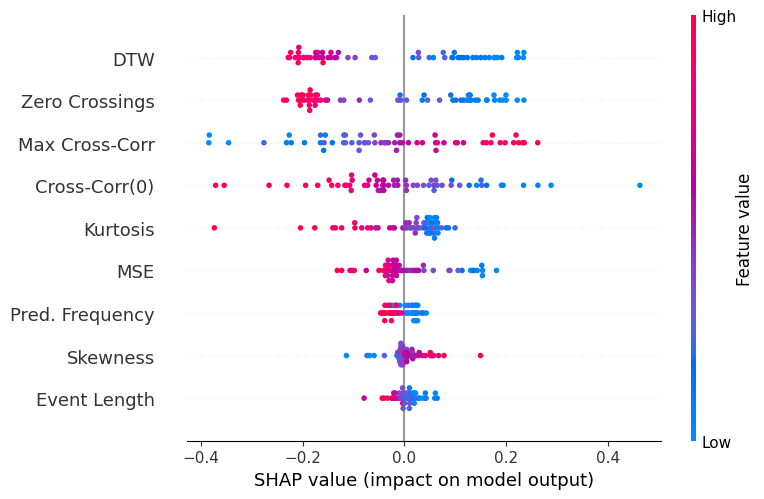}
 \caption{\textbf{Summery plot for Train\_Case\_1}}
 \label{fig:XAi_TrainCase1}
\end{figure}

\begin{figure}[!t]
 \centering
 \includegraphics[width=0.45\textwidth]{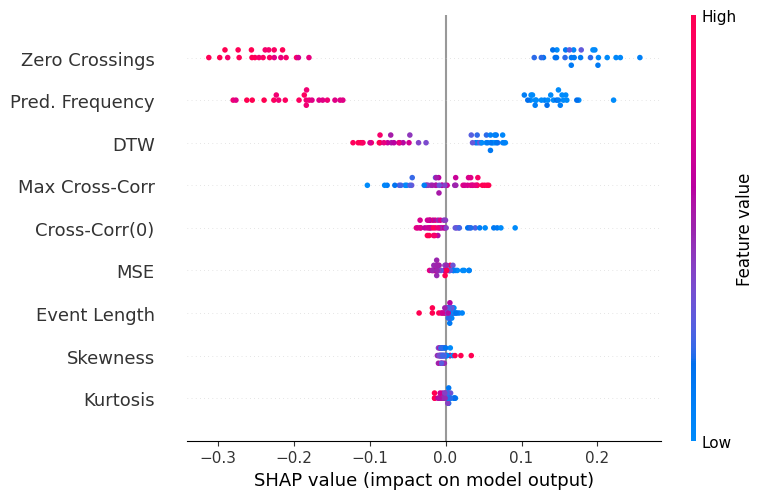}
 \caption{\textbf{Summery plot for Train\_Case\_2}}
 \label{fig:XAi_TrainCase2}
\end{figure}

Considering the high accuracy achieved during validation on the training dataset and the observed performance variations in natural environments, XAI with SHAP was utilized to analyze the contribution of each feature to the referenced ANN model. The summary plot for \texttt{Train\_Case\_1} is shown in  Fig. \ref{fig:XAi_TrainCase1}, while that for \texttt{Train\_Case\_2} is presented in Fig. \ref{fig:XAi_TrainCase2}.

\begin{figure}[!t]
 \centering
 \includegraphics[width=0.45\textwidth]{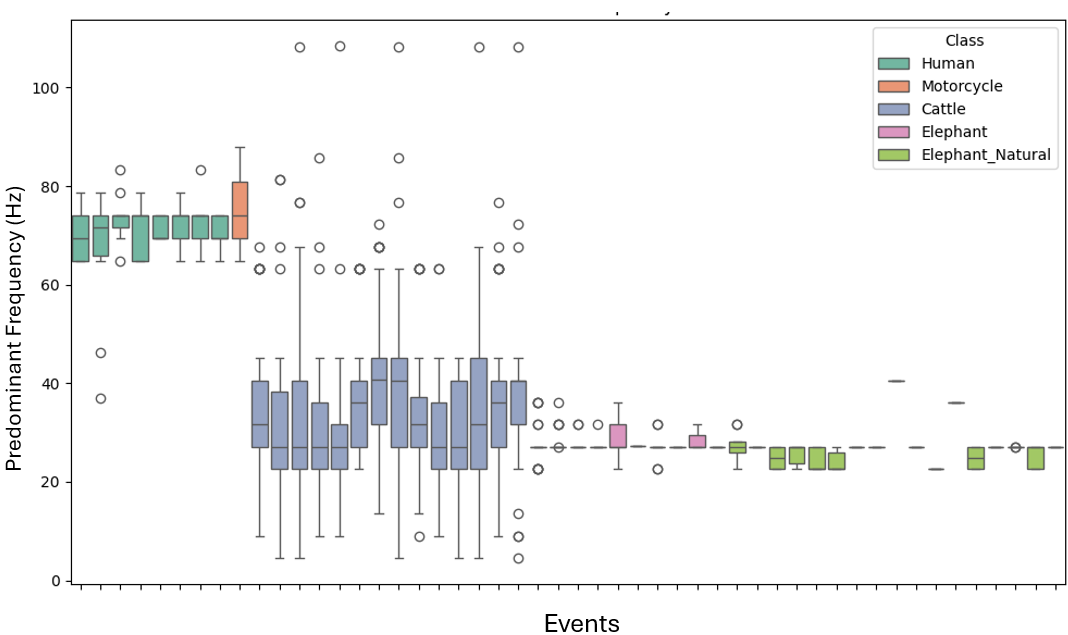}
 \caption{\textbf{Distribution of predominant frequency for each event}}
 \label{fig:BoxPlot_predominentFrequncy}
\end{figure}

As shown in Fig.\ref{fig:XAi_TrainCase1} and Fig. \ref{fig:XAi_TrainCase2}, the number of zero crossings, DTW alignment cost exhibit a significant impact on classification. Morover, Predominant Frequency; a widely used parameter with high significance in related studies \cite{wood2005using, wijayaraja2024towards} demonstrated strong influence under relatively controlled settings (Fig. \ref{fig:XAi_TrainCase2}) but showed a notably reduced impact when trained under more natural conditions (Fig. \ref{fig:XAi_TrainCase1}).

This reduction in the impact of Predominant Frequency may be attributed to the influence of non-elephant events that fall within a similar frequency range to elephant footfalls due to local soil conditions and environmental phenomena. \hl{This effect was further analyzed through the distribution of Predominant Frequency for both elephant and non-elephant events, as illustrated in Fig. \mbox{\ref{fig:BoxPlot_predominentFrequncy}}. It} shows that the Predominant Frequency of cattle footfalls sometimes aligns with the predominant frequency range of elephant footfalls, which is suspected to occur due to wet soil conditions. However, further research is necessary to confirm whether these conditions are specific to Sri Lankan HEC-prone locations, such as paddy fields. 

Despite these challenging conditions, features such as the number of zero crossings and DTW alignment cost maintain high classification performance across different environments. Interestingly, features like event length and skewness have a negligible influence on classification. Thus, removing these features can lower computational costs without significantly affecting the overall accuracy.

\section{Conclusion}
\label{sec:Conclusion}
This study presented a framework to automate the manual classification process in elephant footfall detection and classification. Between the event detection methods compared in this study, STA/LTA method was identified as the most suitable for resource-constrained implementations, incurring the CCW: the novel method presented in this study, tailored specifically for elephant footfall detection requires future attention. However, even at its early stage, the CCW method demonstrated competitive performance with the well-established STA/LTA approach and effectively mitigated the event-merging effect.

The results show that elephant footfalls from 156.6 m away were detected and classified under controlled conditions, whereas, footfalls were detected up to 140 m away in natural environments. The model achieved a footfall detection accuracy of 99\% in controlled settings and 73\% in natural environments, which is among the highest reported in comparable studies. Yet there is a 3\% accuracy drop in the most challenging environments: human habitats prone to HEC. However, the overall performance of the framework in terms of accuracy, range, cost-effectiveness, and practical implementation, concludes that it is a promising solution for long-range elephant detection. Additionally, it was identified that there is an impact on relatively simple features such as DTW alignment cost and number of zero-crossings for classification. Importantly, DTW alignment cost has not been previously explored in related research, underscoring its novelty and potential value.

In future, the CCW method will also be extended to operate beyond elephant footfall detection, offering potential benefits for various ambient intelligence applications. Furthermore, the instrumentation system will be enhanced with automatic gain adjustment to improve accuracy across a broader range of detections.

\section*{Acknowledgment}
The authors express their profound gratitude to Mr. Mendis Wickremasinghe for his invaluable guidance throughout this research. Sincere appreciation is also extended to Major General Lal Gunasekara for his strategic planning and facilitation of field studies. Furthermore, the authors acknowledge the Department of Wildlife Conservation, Sri Lanka and the Department of National Zoological Gardens for their support, guidance, and for granting of necessary permissions to conduct field studies within their premises.

\bibliographystyle{unsrt}
\begin{CJK*}{UTF8}{gbsn}
\bibliography{bibliography}
\end{CJK*}

\begin{IEEEbiography}[{\includegraphics[width=1in,height=1.25in,clip,keepaspectratio]{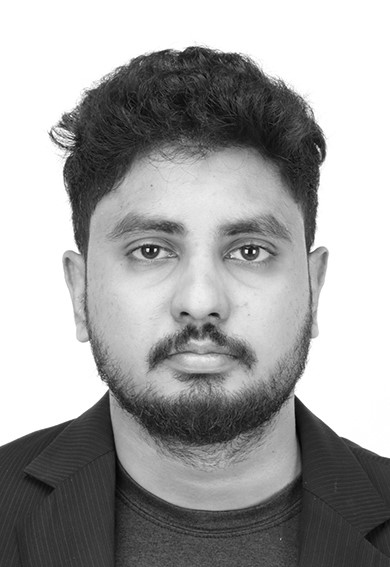}}]{Jaliya L. Wijayaraja} received his BSc Engineering degree with Honors in Electrical and Electronic Engineering from the Sri Lanka Institute of Information Technology (SLIIT), Sri Lanka in 2020. Currently, he is pursuing his MPhil in Computer Networks from SLIIT.

Since 2021, Jaliya has been a research assistant in the Department of Computer Systems Engineering at SLIIT. His research interests include electronic design, signal processing, IoT systems.
\end{IEEEbiography}

\begin{IEEEbiography}[{\includegraphics[width=1in,height=1.25in,clip,keepaspectratio]{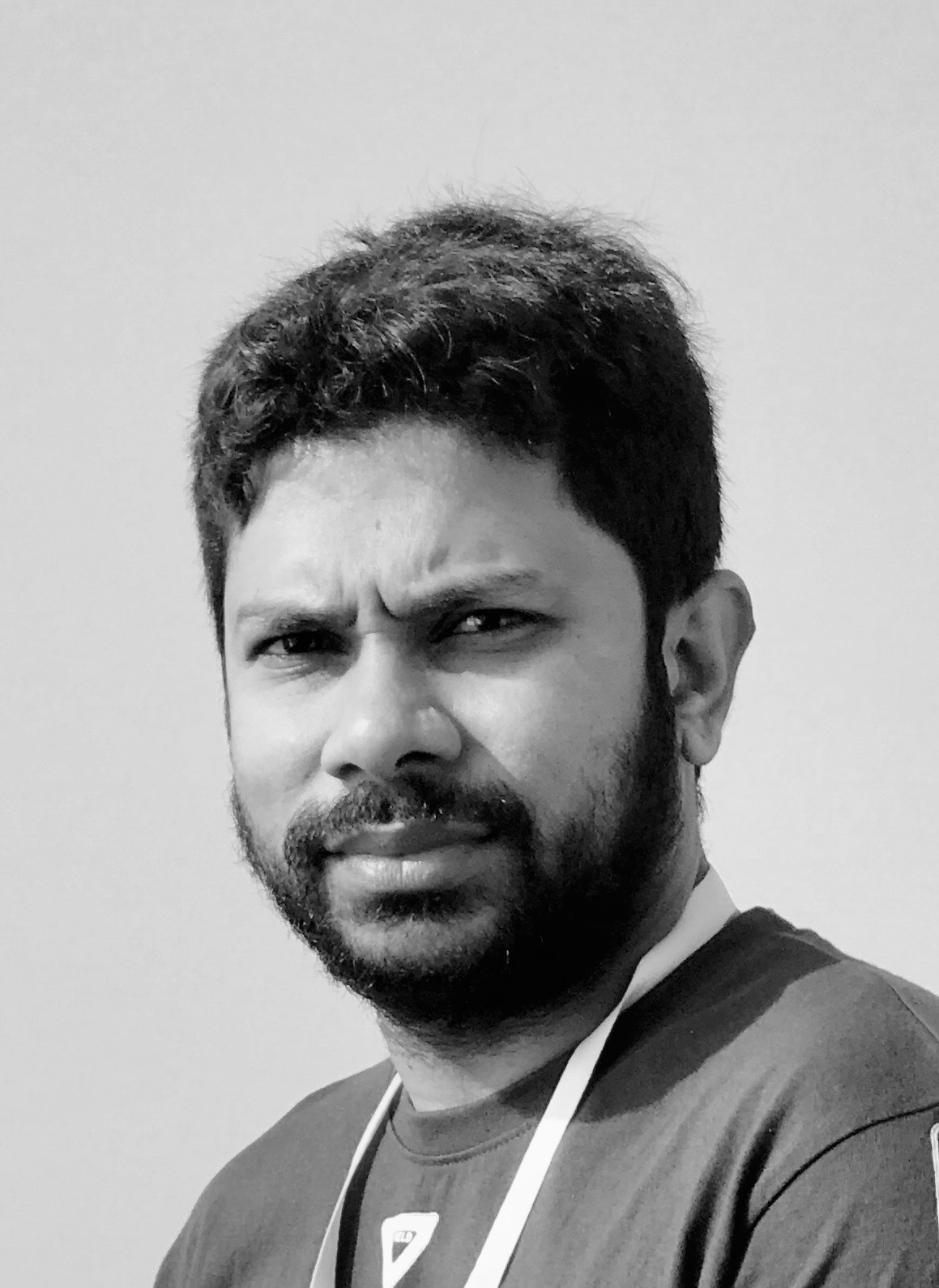}}]{Janaka L. Wijekoon} earned his B.Sc. degree from SLIIT, Sri Lanka, in 2010, and went on to complete his M.Sc. (2013) and Ph.D. (2016) degrees at Keio University, Japan. With a background as a postdoctoral researcher with extensive experience in academia, his expertise centres around leveraging AIoT to advance smart societies. His research focuses on collaborative smart infrastructural design and implementations for secure and efficient resource allocation in smart societies and protocol design for data exchange among stakeholders to improve agricultural productivity. 
\end{IEEEbiography}

\begin{IEEEbiography}[{\includegraphics[width=1in,height=1.25in,clip,keepaspectratio]{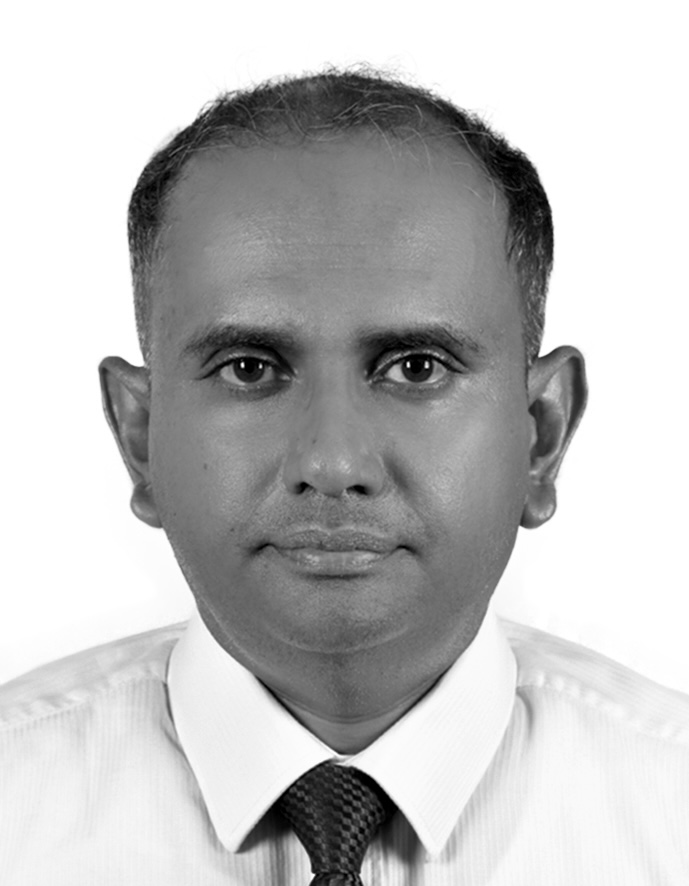}}]{Malitha Wijesundara} earned his BEng degree with Honors in Electronic Engineering from the University of Warwick, United Kingdom in 1998. He then went on to complete his PhD in Computer Engineering at the National University of Singapore. He has extensive experience in Embedded Systems, Wireless Sensor Networks, Media Streaming for E-Learning and Energy Harvesting for Elephant Tracking. His research focuses on developing technology solutions to challenges faced by rural and underprivileged communities. He is a member of IEEE, ISACA and CS(SL).
\end{IEEEbiography}

\EOD
\end{document}